\documentclass[10pt,journal,compsoc]{IEEEtran}

\usepackage[english]{babel}
\usepackage[utf8]{inputenc}

\usepackage[dvipsnames]{xcolor}

\usepackage{booktabs} %
\usepackage{multirow}
\usepackage{multicol}
\usepackage{colortbl} %

\usepackage{graphicx}
\usepackage{amsmath}
\usepackage{amssymb}
\usepackage{comment}

\usepackage{footmisc}

\usepackage{ulem} %

\PassOptionsToPackage{unicode}{hyperref}
\PassOptionsToPackage{naturalnames}{hyperref}
\usepackage{placeins} %

\newcommand{\parag}[1]{\textbf{#1}}

\newif\ifdraft

\definecolor{color_table_highlight}{gray}{0.85}
\newcolumntype{h}{>{\columncolor{color_table_highlight}}c}

\newcommand{\segmetricsA}[1]{
	\multicolumn{2}{c}{Pixel-level} 
	& \multicolumn{3}{#1}{Segment-level}
}

\newcommand{\segmetricsB}{
	AP $\uparrow$ 
	& FPR$_{95}$ $\downarrow$ 
	& $\overline{\mbox{sIoU}}\uparrow$ 
	& $\overline{\mbox{PPV}}\uparrow$ 
	& $\overline{F_1}\uparrow$
}

\usepackage[pagebackref=true,breaklinks=true,colorlinks]{hyperref}
\hypersetup{
	pdftitle={Detecting Road Obstacles by Erasing Them},
	pdfauthor={Krzysztof Lis, Sina Honari, Pascal Fua, Mathieu Salzmann},
	hyperfootnotes = true,
	pdfcreator={},
	pdfproducer={},
}
\usepackage{cleveref}

\newif \ifrelease
\releasetrue

\begin{document}

\title{Detecting Road Obstacles by Erasing Them}
\author{Krzysztof~Lis,
	Sina~Honari,
	Pascal~Fua,
	Mathieu~Salzmann%
\IEEEcompsocitemizethanks{
\IEEEcompsocthanksitem{
All authors are affiliated with the Computer Vision Laboratory, EPFL.
lis.krzysztof@protonmail.com, sina.honari@gmail.com, mathieu.salzmann@epfl.ch, pascal.fua@epfl.ch
}
\IEEEcompsocthanksitem{
This work was supported in part by the International Chair Drive for All - MINES ParisTech - Peugeot-Citroën - Safran - Valeo.
}
}}

\IEEEtitleabstractindextext{
\begin{abstract}
  
Vehicles can encounter a myriad of obstacles on the road, and it is impossible to record them all beforehand to train a detector. Instead, we select image patches and inpaint them with the surrounding road texture, which tends to remove obstacles from those patches. 
We then use a network trained to recognize discrepancies between the original patch and the inpainted one, which signals an erased obstacle.  

\end{abstract}

}

\maketitle
\IEEEdisplaynontitleabstractindextext	
\IEEEpeerreviewmaketitle

\section{Introduction}\label{sec:intro}

Modern methods provide a nearly complete toolkit for vision-based autonomous driving. That includes road segmentation~\cite{Chen17rbnet,Munoz17deep}, lane-finding~\cite{Liu21end,Qin20a}, vehicle and pedestrian detection~\cite{Ren15,Dollar12b,Chabot17manta}, and multi-class semantic~\cite{Zhao17b,Chen17a,Romera17}, instance~\cite{He17a} and panoptic~\cite{Kirillov19,Xiong19b} segmentation.
Moreover, vision-based assistance is now widely accepted in the market~\cite{LaneAssist} and enables vehicles to plan a path within the predicted drivable space while avoiding other traffic participants. However, the risk of missing strange and unexpected obstacles lying on the road remains. Such obstacles are as rare as they are diverse, which prevents direct application of the now standard approach of training deep networks by showing them an exhaustive set of annotated samples.

\begin{figure} 
\centering
	\includegraphics[width=0.49\linewidth]{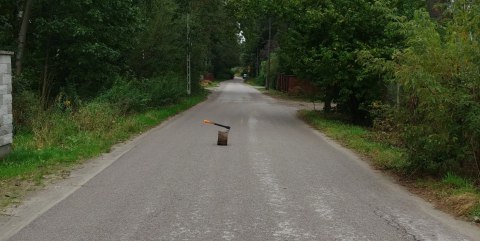}
	\includegraphics[width=0.49\linewidth]{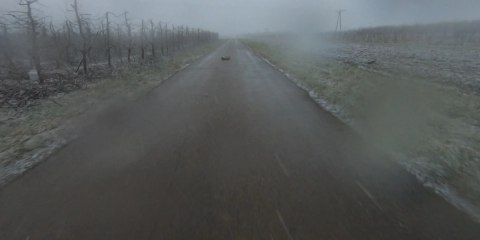}
	\includegraphics[width=0.49\linewidth]{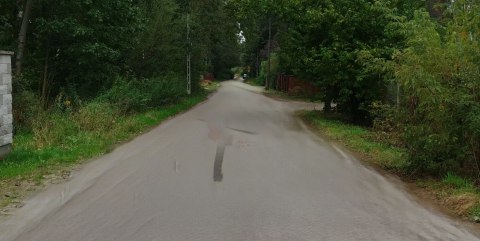}
	\includegraphics[width=0.49\linewidth]{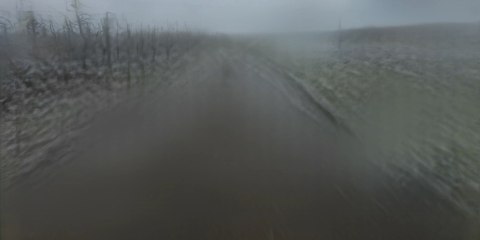}
	\includegraphics[width=0.49\linewidth]{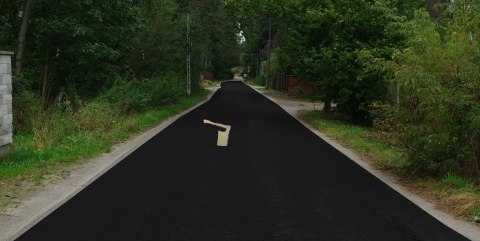}
	\includegraphics[width=0.49\linewidth]{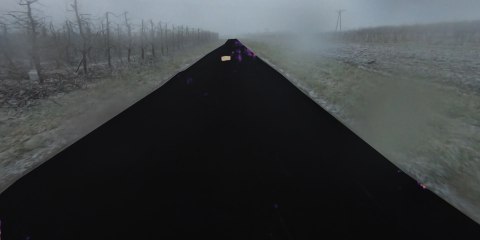}
\caption{\small{\bf Detecting unexpected obstacles in good and bad weather.}{\bf Top:} Objects  one would not expect to see on a road and that are not featured in standard databases. 
	{\bf  Middle:} The road area has been inpainted.
	{\bf Bottom:}  After comparing the original and inpainted images, our discrepancy network returns a binary mask that denotes the obstacle locations.
}
\label{fig:teaser}
\end{figure}

In practice, detecting such unexpected obstacles often requires LiDAR sensors~\cite{SinghKamireddypalli20} or multiple cameras~\cite{Pinggera16}. Here, we propose instead a method that only needs
a single RGB image to detect obstacles in the drivable area, under the assumption that objects outside that area are irrelevant because a self-driving car will detect the road before planning to drive and will not leave the drivable area of its own accord. To demonstrate this to be a viable assumption, we will show results given either the ground-truth location of the road edges or only an imperfect road segmentation produced by an off-the-shelf segmentation algorithm.

Our approach relies on the fact that obstacles look different from the surrounding road surface. 
We thus detect them by inpainting image-patches using their surroundings and then checking how similar the inpainted patch is to the original one. 
While a similar intuition has been used to detect anomalies in several application scenarios, such as detecting manufacturing defects~\cite{Zavrtanik20,Haselmann18} or anomalous faces~\cite{Bhattad18}, the very constrained nature of these tasks made it possible to rely on simple comparisons of handcrafted features. By contrast, on roads, this would yield many false positives due to road markings, diversity in road texture, and obstacles extending beyond the inpainted patch.

Our solution is to introduce a {\it discrepancy network} trained to recognize which differences between the inpainted patch and the original one are significant. It returns a per-pixel heatmap denoting the presence of obstacles. To train it to handle objects that are {\it not} part of the training database,  we generate samples featuring synthetic obstacles by moving existing training objects, such as road signs and people, onto the road.

Our experiments show that our discrepancy network trained solely on Cityscapes~\cite{Cordts16} objects successfully detects obstacles on images depicting significantly different road scenes, without requiring any annotated data nor any re-training for these new scenes. In other words, our method generalizes well to previously unseen real obstacles and new road surfaces. It outperforms earlier monocular road anomaly detectors~\cite{Lis19,Blum19,Malinin18,Bevandic19} on the {\it Lost \& Found}~\cite{Pinggera16} data featured in the {\it Fishyscapes} benchmark~\cite{Blum19}, as well as on our own newly collected dataset featuring additional unusual objects and road surfaces.

Our contribution is therefore a simple but effective approach to detecting obstacles that never appeared in any training database, given only a single RGB image. We also contribute a new dataset for evaluating anomaly detection models.
\ifrelease %
We will make our obstacle detection code\footref{fn:repo}, our semi-synthetic training dataset, and our new benchmark dataset\footref{fn:dataset} publicly available. 

\footnotetext[1]{\label{fn:repo}
Code and semi-synthetic training set:\\ \href{https://github.com/cvlab-epfl/erasing-road-obstacles}{github.com/cvlab-epfl/erasing-road-obstacles}
}
\footnotetext[2]{\label{fn:dataset}{\it Road Obstacles} dataset:\\ \href{https://www.epfl.ch/labs/cvlab/data/road-obstacles-2021/}{www.epfl.ch/labs/cvlab/data/road-obstacles-2021}
}
\else
We will make our obstacle detection code, our semi-synthetic training dataset, and our new benchmark dataset publicly available. 
\fi

\section{Related Work}\label{sec:related}

In this section, we briefly review methods that, as ours, rely on a single RGB image for obstacle detection and do not use explicit obstacle training sets. For a more complete survey of obstacle detection algorithms, we refer the reader to~\cite{Pinggera16,Ramos17,Gupta18,Xue19}. 

\subsection{Image Reconstruction for Anomaly Detection}

If an image is reconstructed so as to preserve the 
appearance in normal regions and discard anomalous ones,
those anomalies can be detected by comparing the input to the reconstruction.
This has been achieved in several ways we now discuss.

In~\cite{Akcay18}, the distribution of inlier images is modeled by training an autoencoder,
followed by a feature comparison between the input and reconstruction. %
Auto-encoding has been used to find highway obstacles in~\cite{Creusot15, Munawar17},
where input patches are passed through a Restricted Boltzmann Machine.
Its limited expressive power is supposed to preserve the smooth road surface while altering obstacles.
This approach, however, does not address textured road surfaces.
As opposed to encoding the input image, the method of~\cite{Schlegl17} trains a generator to capture the training distribution
and searches for a latent vector producing an image most similar to the input.
However, this method operates on microscopic scans of tissue samples, and, to our knowledge, it has not been applied to data with distributions as diverse as outdoor road scenes.

Other methods~\cite{Lis19,Xia20} explicitly restrict the intermediate representation of the scene to a dense semantic map,
and synthesize a plausible matching image using conditional GANs for image translation~\cite{Isola16,Wang18c}.
Since the anomalous regions are not represented by the typical semantic classes,
their appearance will be altered by this process.
In~\cite{Lis19}, the input and synthesized images are compared using a learned discrepancy module,
while~\cite{Xia20} uses a feature distance measure. %
JSRNet \cite{Vojir21} takes a similar approach, but uses the activations of the last layer as the bottleneck instead of a semantic map.
A reconstruction network then recreates the image that is compared to the original using a SSIM measure.

Rather than encoding the input, one can remove parts of the image and inpaint them based on the surrounding context.
In~\cite{Haselmann18} square patches are inpainted and compared with an $L_{1}$ metric to detect material defects;
the method of~\cite{Zavrtanik20} combines the reconstructions obtained with a set of random inpainting masks
and uses a multi-scale gradient magnitude similarity metric for comparison.
In the context of road scenes, \cite{Munawar15}~proposes to compare the road appearance to similar images memorized from previous video frames;
however this would lead to false positives when entering an area with a new road texture.
Furthermore, these methods assume high fidelity of the reconstruction, and detect every visible difference as an anomaly.
In outdoor scenes with road markings and diverse surface textures, the inpainting is bound to be imperfect. We address this by training a discrepancy network to focus on the relevant differences.

Methods specifically designed for removing dynamic objects from traffic scenes have been presented in~\cite{Bescos18,Berlincioni19}, but they rely on object masks being known {\it a-priori} without having to detect them first.

\subsection{Anomaly Detection in Semantic Segmentation}

The problem of detecting anomalies can be posed as one of open-set semantic segmentation. With standard, fully-supervised semantic segmentation networks, all pixels, including the anomalous ones, will be classified into one of the training semantic categories. Open-set semantic segmentation then aims to find the outliers in the resulting semantic maps.
The method of~\cite{Bevandic19} introduces an outlier detection head sharing backbone features with the semantic segmentation one.
It is trained using extensive out-of-distribution data, injecting outlier patches drawn from ImageNet-1k~\cite{Deng09} into the Cityscapes and Vistas~\cite{Neuhold17} scenes.

The work of~\cite{Blum19} proposes to learn the inlier distribution of features extracted from a layer of the semantic segmentation network.
It uses a normalizing flow to bijectively map the features to latent vectors following a Gaussian distribution.
The mapping is trained to maximize the likelihood of the features observed in inlier samples.

Another approach involves estimating the uncertainty of label prediction based on the assumption that outlier regions should yield low classification confidence.
Bayesian deep learning treats the network weights and outputs as probability distributions.
In practice, dropout~\cite{Srivastava14} is used to approximate the distribution over model weights,
applied to semantic segmentation in~\cite{Kendall15b,Kendall17}. %
Alternatively an ensemble of networks can be used~\cite{Lakshminarayanan17, Gustafsson20}.
In~\cite{Malinin18}, the Dirichlet differential entropy is used as a measure of uncertainty.
We find that the performance of these statistical methods as obstacle detectors degrades significantly 
when faced with road surfaces differing from the training set,
as the novel textures are treated as anomalies.
We will show that our approach based on comparing the original image with an inpainted one with a discrepancy detector does not suffer from this drawback.

In~\cite{Chan21a}, generalization is achieved by training the segmentation network to maximize the output entropy on explicit out-of-distribution samples. This is done by injecting COCO~\cite{Lin14a} objects into Cityscapes frames. According to our experiments this is effective for some types of obstacles, presumably ones more similar to COCO objects, but not all of them.
The additional samples can also be created by a generative model jointly optimized with the detector to output points on the edge of the training distribution~\cite{Grcic21}.  A normalizing flow is used to generate synthetic training anomalies in a similar manner in~\cite{Grcic21b}.
We also use synthetic data to train our discrepancy network but our object-pasting scheme is much simpler. Hence, these approaches are complementary to ours, which focuses on the inference mechanism. Hence it, would make sense to combine them in future work.

Finally, the SynBoost~\cite{DiBlase21} method fuses resynthesis dissimilarity with semantic segmentation uncertainty estimates.
This improves upon previous resynthesis-based approaches. However, we will show that our method outperforms SynBoost while requiring neither detailed semantic segmentation nor associated uncertainty; we only need to segment the road area to guide our inpainter.

Another way to detect semantic anomalies is to analyze the distribution of the predicted semantic logits.
A low value of the maximum logit implies an anomaly as none of the classes is confidently predicted \cite{Hendrycks17b}.
However the logit values differ strongly between each class, making a single threshold work poorly.
The method of \cite{Jung21} standardizes the logit values to align distributions between classes.
It also removes spurious detections on object edges by post-processing. 
{In our approach we do not reason about the semantic predictions - we only use the semantic network as a convenient way to segment the road but a simpler road or lane detector could be used instead.}

\section{Approach}\label{sec:method}

Our goal is to identify obstacles that are objects on the road the vehicle could collide with.  This means that they are within the space deemed drivable by a previous stage in the self-driving perception pipeline.  As such, they are the most relevant obstacles as a competent driving system will only plan trajectories within that space.

In other words, our chosen task is to identify all pixels, within that estimated road area, that denote obstacles. This is difficult because obstacles can take many forms. 
Furthermore, because they are unexpected, there is no guarantee that they were present in the database used to train a network to recognize them. 
Hence, the network must be made to respond to {\it objects that does not belong to the road} without any clear description, or even exemplars, of these objects.	

To this end, given a binary mask denoting the drivable area in the image, we propose the following two-step approach:
\begin{enumerate}

	\item Erase the obstacles by removing road patches and inpainting them in a sliding-window manner;
	
	\item Use a discrepancy network to compare the original image to the inpainted one and decide if they are similar enough.  
	
\end{enumerate}
The intuition behind this scheme is that, if there is an obstacle, the  inpainted area will look very different from the original image. However, even if there is no obstacle, the inpainted area will be similar to the original one, but not strictly equal. Hence, the discrepancy network is needed to assess if they are dissimilar enough to flag a potential obstacle. It yields a heatmap denoting the likelihood for each pixel in the drivable area of belonging to an obstacle. In the remainder of this section, we discuss these two steps in more detail. 

\subsection{Drivable Area}

\begin{figure} 
\centering

\includegraphics[width=0.48\linewidth]{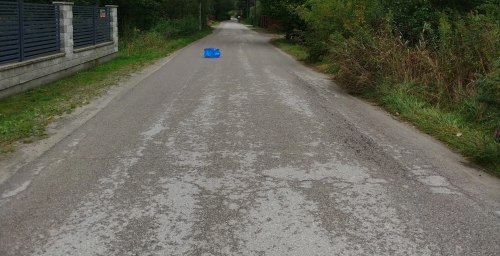}
\includegraphics[width=0.48\linewidth]{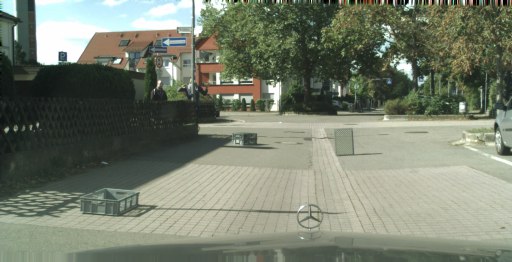}

\includegraphics[width=0.48\linewidth]{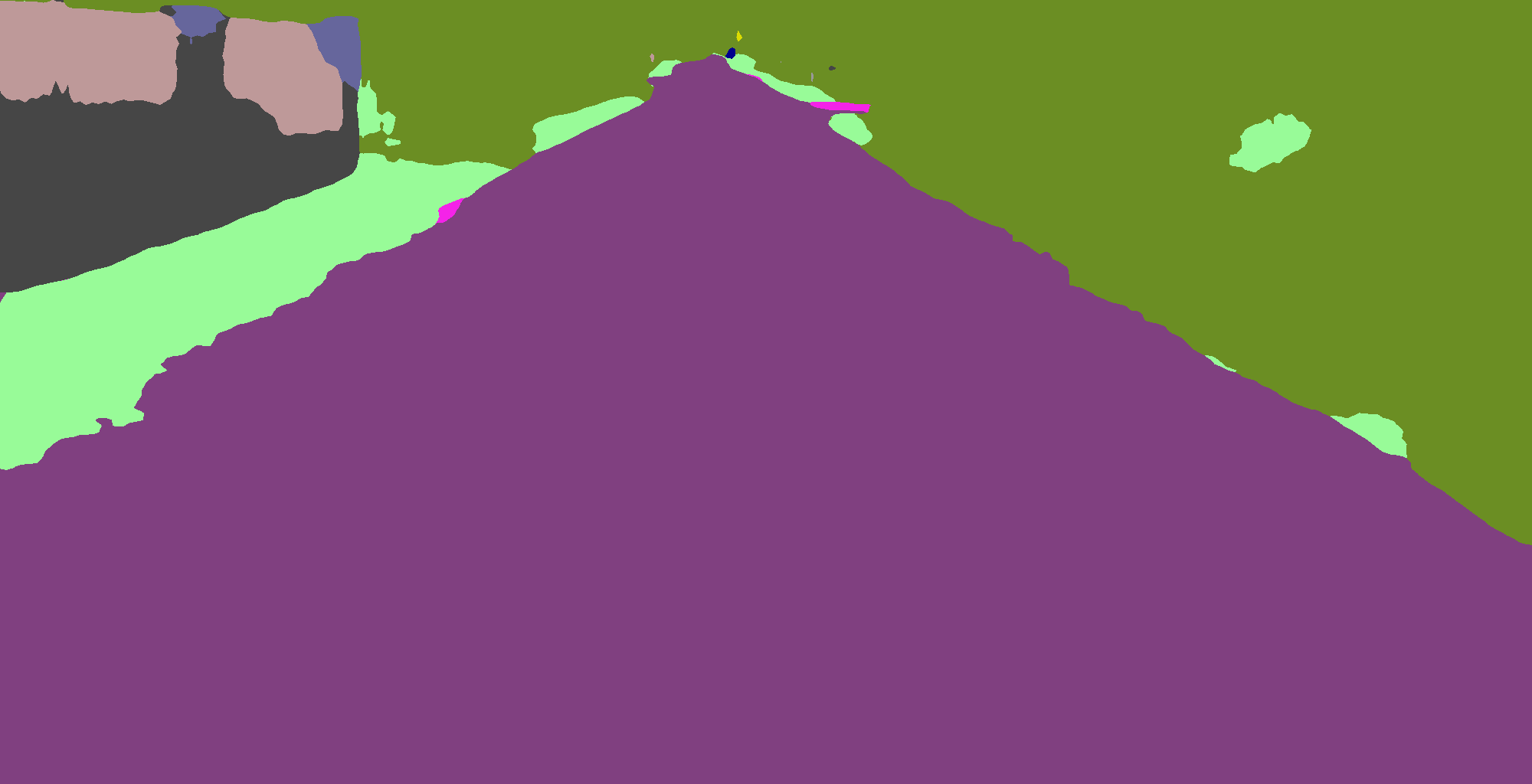}
\includegraphics[width=0.48\linewidth]{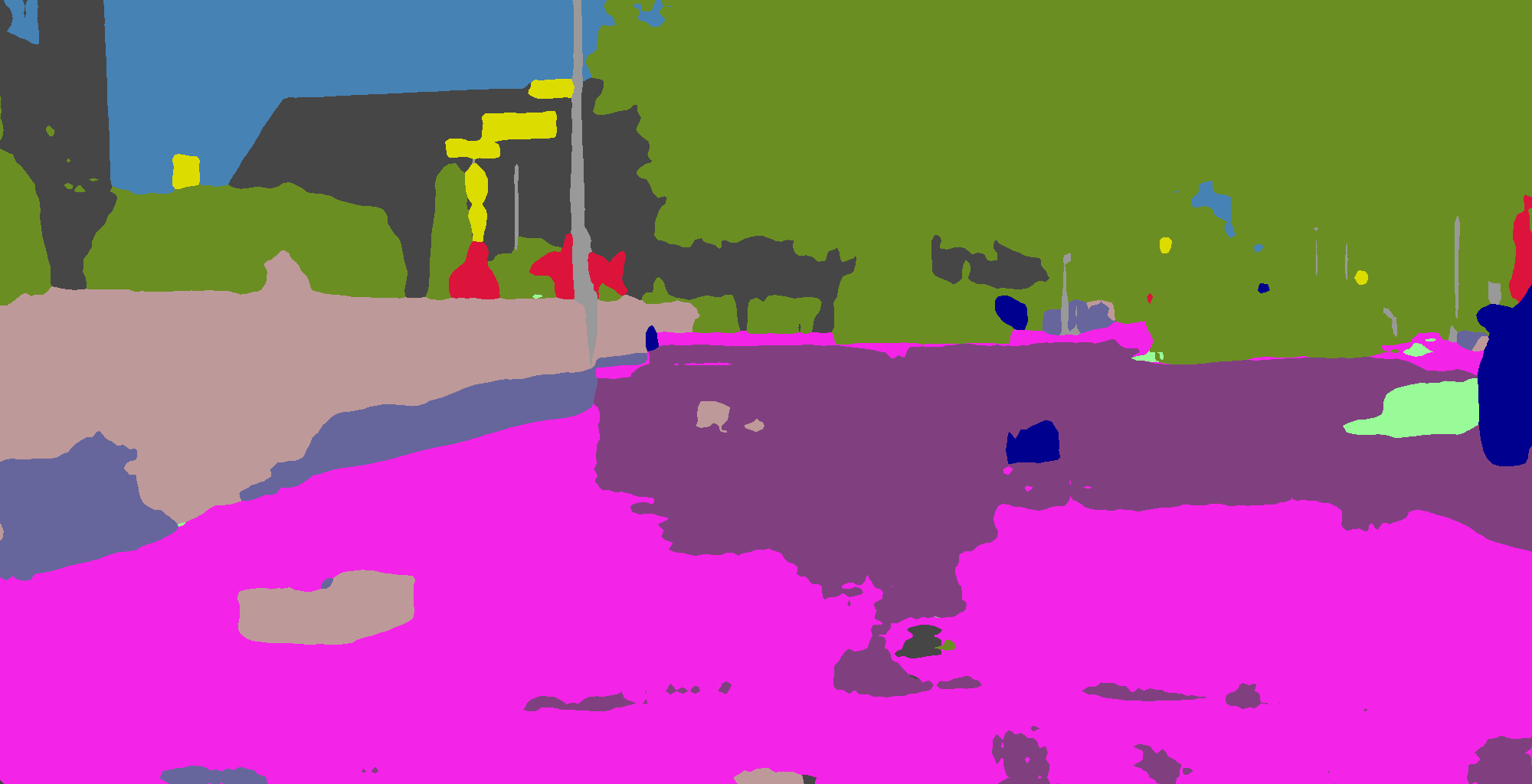}

\includegraphics[width=0.48\linewidth]{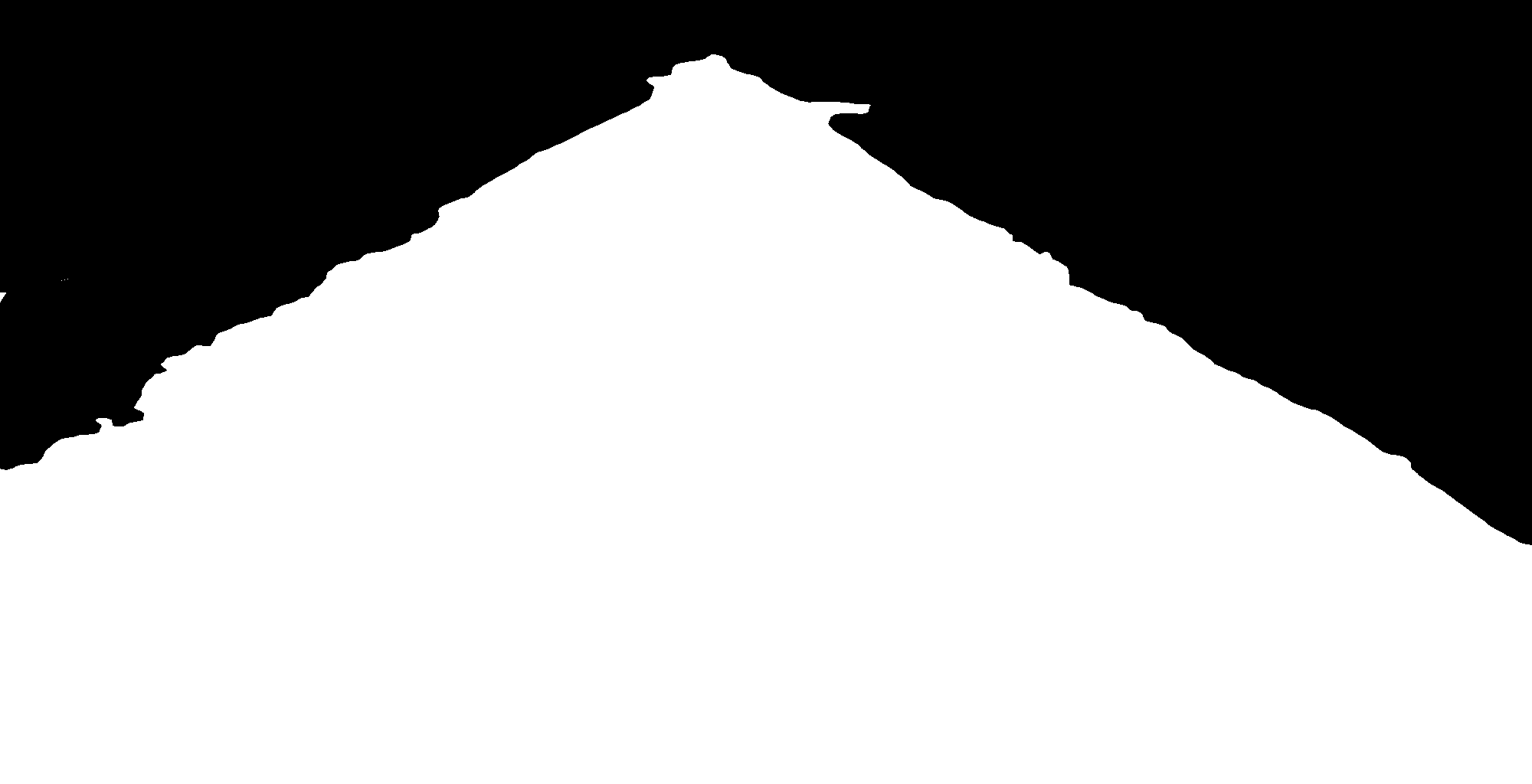}
\includegraphics[width=0.48\linewidth]{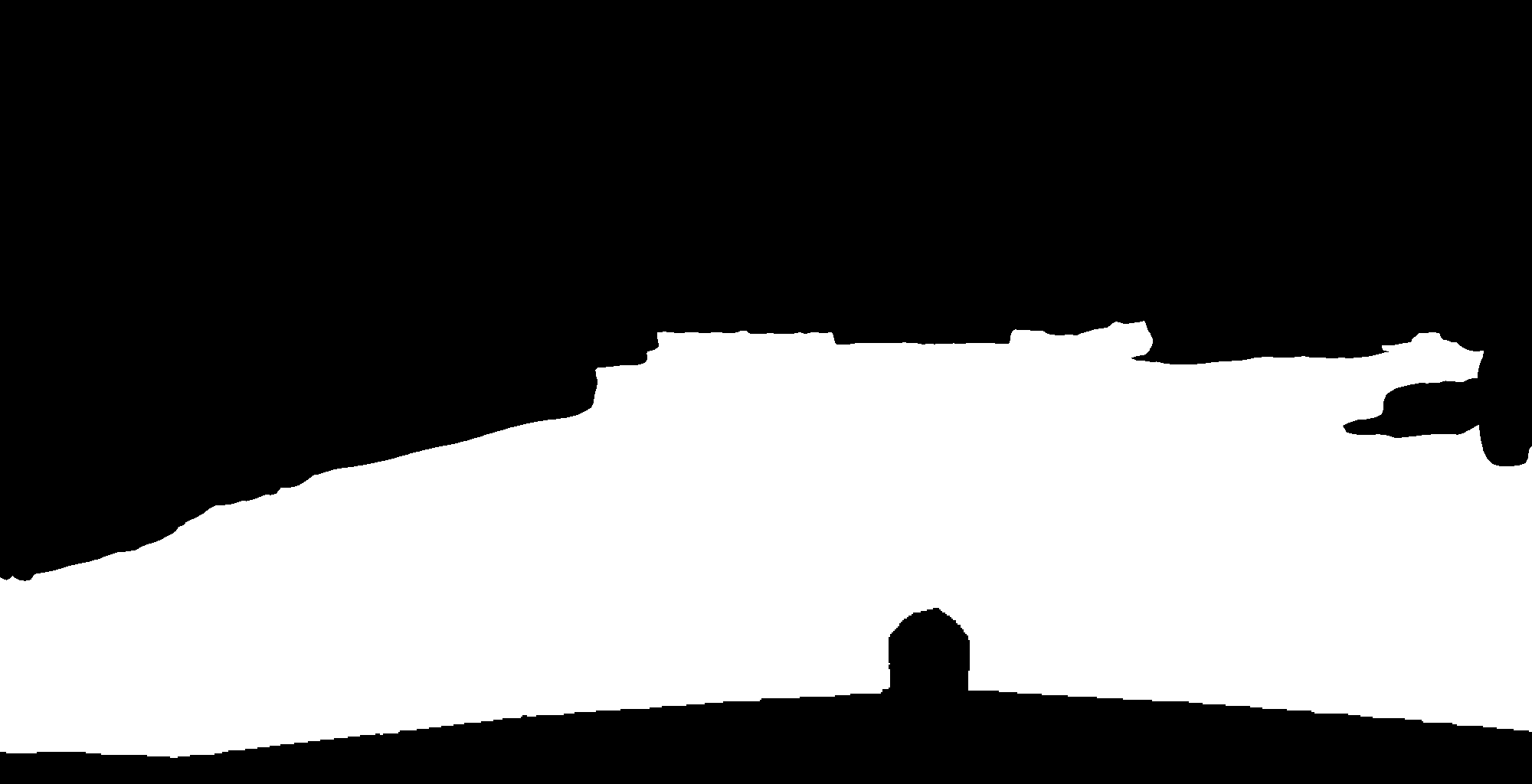}

\caption{\small{\bf Drivable space from semantic segmentation.}{\bf Top:} Input images.
{\bf Middle:} Semantic segmentation performed by PSP-Net~\cite{Zhao17b}, the class colors follow Cityscapes convention.
{\bf Bottom:} We take the drivable space to be the union of {\it road} (purple) and {\it sidewalk} (magenta) pixels.
The parts of obstacles can sometimes be classified as non-road, so we include the regions of other classes fully enclosed within the road area.
In {\it Lost \& Found}, the known ego-vehicle mask is excluded.
}
\label{fig:drivable_space}
\end{figure}

A self-driving system must determine the road area within which it can move. Our method detects obstacles within the area identified as drivable, as these are the only ones that might be on the vehicle's planned path and can endanger it. Of course, a part of the road may be mistakenly marked as non-drivable, causing obstacles it contains to be ignored as shown Fig.~\ref{fig:obstacle_outside_road}. However, this does not compromise safety because the vehicle will never attempt to go there. 

Our approach can exploit any method that delivers the required drivable area information. In our experiments, we use the PSP-Net semantic segmentation network of~\cite{Zhao17b} trained on the Cityscapes dataset~\cite{Cordts16}, as implemented in the framework of~\cite{GluonCV20}. We take the road area to be all pixels classified as either {\it road} or {\it sidewalk}, since the many road textures we are targeting can be classified as either. Note that standard categories, such as {\it car} and {\it pedestrian}, are inherently accounted for by PSP-Net. Hence,  we focus on the unusual obstacles for which no training data, either supervised not unsupervised, is available. Nevertheless, since such unusual obstacles could be partially classified as non-road, we include the regions containing other classes that are fully enclosed within the road area. Fig.~\ref{fig:drivable_space} demonstrates this process.  As a limit case, we also evaluate the case of perfect road detection by using the ground truth road mask.

Our approach needs only a coarse mask of the drivable area, so in a practical deployment the semantic segmentation
can be replaced by computationally more efficient system, for instance predicting just the road edges and filling the space between them.

\subsection{Inpainting}\label{sec:method_inpainting}

\begin{figure} 
\centering
	\includegraphics[width=1.00\linewidth]{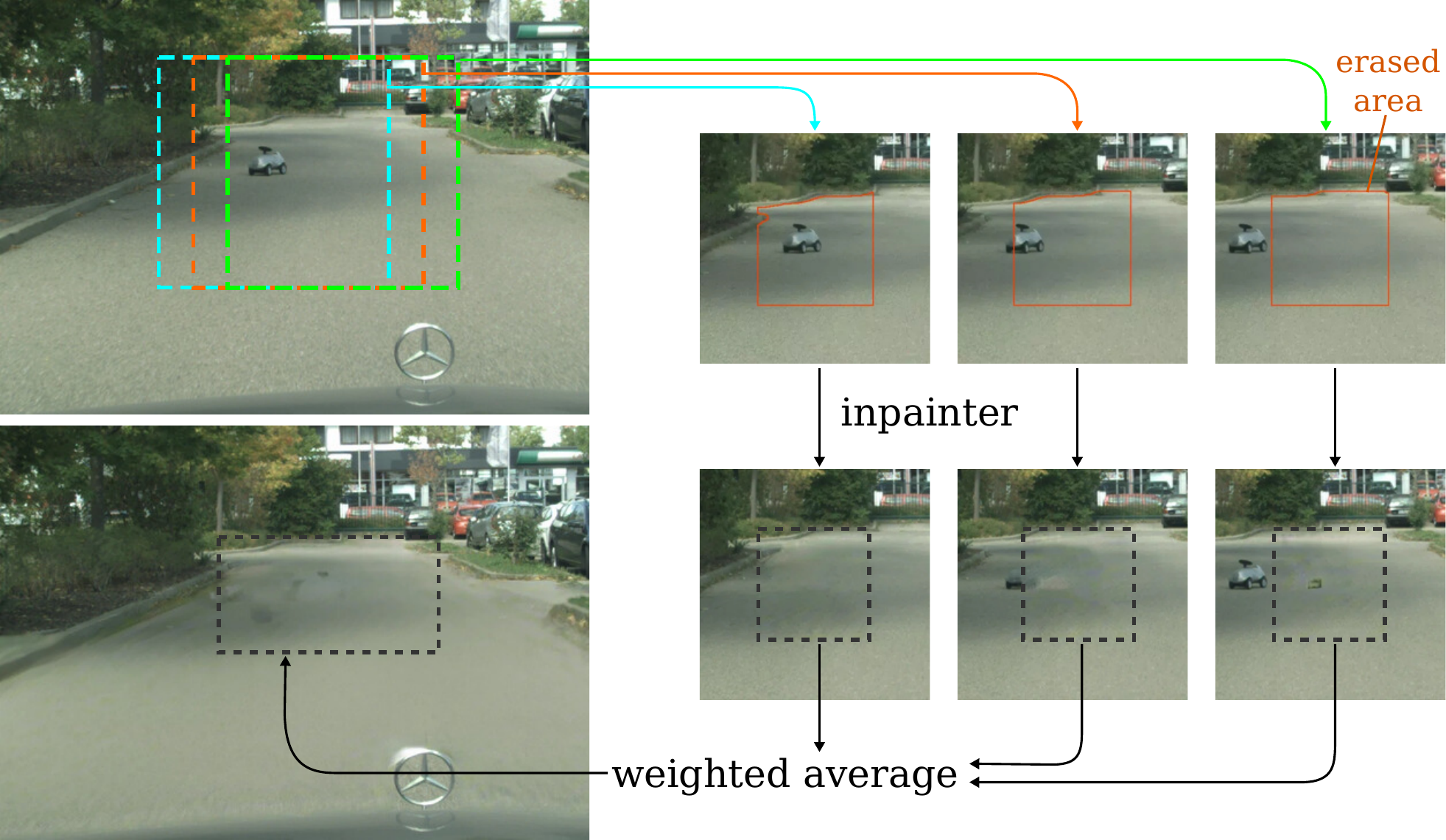}
\caption{{\bf Sliding window inpainting of the road surface.}
We extract $400 \times 400$ context patches then erase and inpaint the road area contained within the central $200 \times 200$ of the patch.
Finally we fuse the inpaintings which reconstructs the road appearance while removing localized obstacles.
Note how the process was able to preserve the shadow of the trees.
}
\label{fig:method_inpainting_pipeline}
\end{figure}

To erase the obstacles while preserving the surrounding road appearance, we use a general-purpose inpainter~\cite{Yu19} that relies on an adversarial approach to ensure that the inpainted image looks realistic. We use a version of this model trained on the scene recognition dataset Places2~\cite{Zhou17h}, and do not train it further.

The inpainter is given an image in which a part has been replaced by black pixels and the pixel mask of the removed area. It outputs an image with the missing part filled in so as to best maintain the continuity and content of the scene.

A naive way to use it would be to inpaint the entire road area at once. This, however, would provide no indication to the inpainter of the road appearance, leading to inpainted images that differ from the original ones in the whole road region, thus precluding subsequent obstacle detection. Instead, we inpaint road patches to provide sufficient context for the network to reconstruct the road surface, as shown in Fig.~\ref{fig:method_inpainting_pipeline}. The patches nonetheless need to be large enough to encompass obstacles whose size we do not know {\it a priori}. We therefore follow a sliding-window approach, inpainting patches of $200 \times 200$ pixels of drivable area within $400 \times 400$ image regions to provide context.

While an obstacle is usually nicely erased when the area to inpaint encloses it completely, the inpainter is able to re-create the obstacles that are only partially contained in the  inpainted region. To resolve this, we use consecutive patches with a relative overlap of 0.7, increasing the likelihood of having at least one patch that covers the entire obstacle. This means that each image pixel is inpainted multiple times. We then fuse the multiple inpaintings of each pixel by weighted averaging, where the weight of each inpainting is computed based on the Manhattan distance between the corresponding patch center and the pixel location of interest. 
Formally, a patch centered at location ${\bf c}_j = [u_j, v_j]^{\top}$ contributes to the inpainting of a pixel at location ${\bf p} = [u,v]^{\top}$ with a weight 
\begin{equation}
w = \frac{
	1 - \frac{2}{s}\max(|u - u_j|, |v - v_j|)
}{
	\sum_{[u_i, v_i]\in \Pi(u, v)} 1 - \frac{2}{s} \max(|u - u_i|, |v - v_i|)
}
\; ,
\end{equation}
where $s$ is the patch width or height
and $\Pi(u, v)$ is the set of patches overlapping the $[u,v]^{\top}$ pixel, with patch $i$ centered at $[u_i, v_i]$.

Our inpainting strategy enables us to generalize to new road surfaces, such as those of our Road Obstacles dataset. By contrast, the GAN synthesizers of \cite{Lis19,Xia20,DiBlase21} always produce images resembling the Cityscapes training set.

\subsection{Discrepancy Network}\label{sec:method_discrepancy}

\begin{figure} 
\centering
	\includegraphics[width=1.00\linewidth]{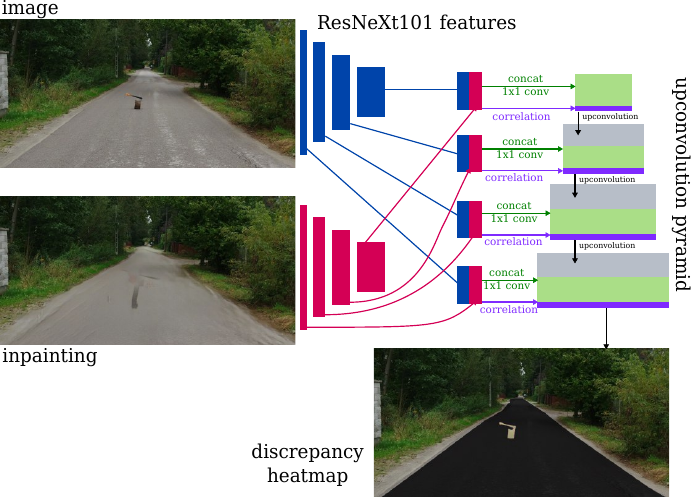}
\caption{{\bf Architecture of the discrepancy network}
}
\label{fig:method_discrepancy_architecture}
\end{figure}

While our inpainting strategy preserves the general appearance of the road surface, it still yields unavoidable imperfections due to road markings, texture details, and the non-zero contributions of obstacles located close to the patch edges. Thus, simply comparing the original image with the inpainted one via pixel difference would yield many false positive detections. To handle this, we introduce a {\it discrepancy network} that we train to distinguish significant differences from inpainting artifacts.

We implement our discrepancy network using a two-stream architecture, shown in Fig.~\ref{fig:method_discrepancy_architecture}, that takes as input the original image and the inpainted one. 
Both inputs are first processed by a ResNeXt101~\cite{Xie17b} feature extractor, pretrained for ImageNet~\cite{Deng09} and frozen at training time.
The resulting features are then concatenated and fused through $1\times1$ convolutions. 
Furthermore, we compute a point-wise correlation map: At each location in the feature map, we calculate the dot product between the image feature vector and the corresponding inpainting feature vector.
We append these dot products as an additional channel to the output of the $1\times1$ convolutions. 
The concatenated features are then passed to an upconvolution pyramid, and we obtain the desired  heatmap via a softmax.

\subsubsection{Training the Discrepancy Network}

\begin{figure} 
\centering
\includegraphics[width=0.48\linewidth]{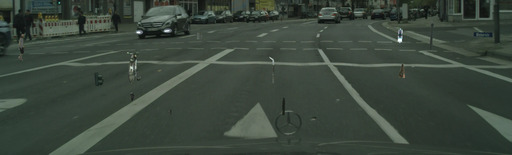}
\includegraphics[width=0.48\linewidth]{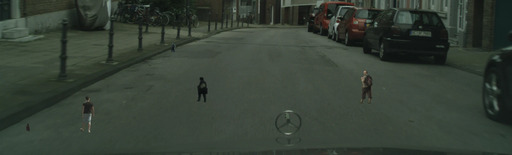}
\includegraphics[width=0.48\linewidth]{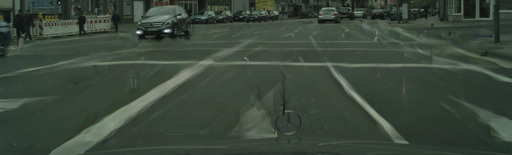}
\includegraphics[width=0.48\linewidth]{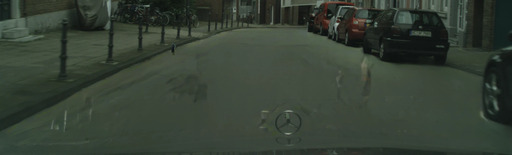}
\includegraphics[width=0.48\linewidth]{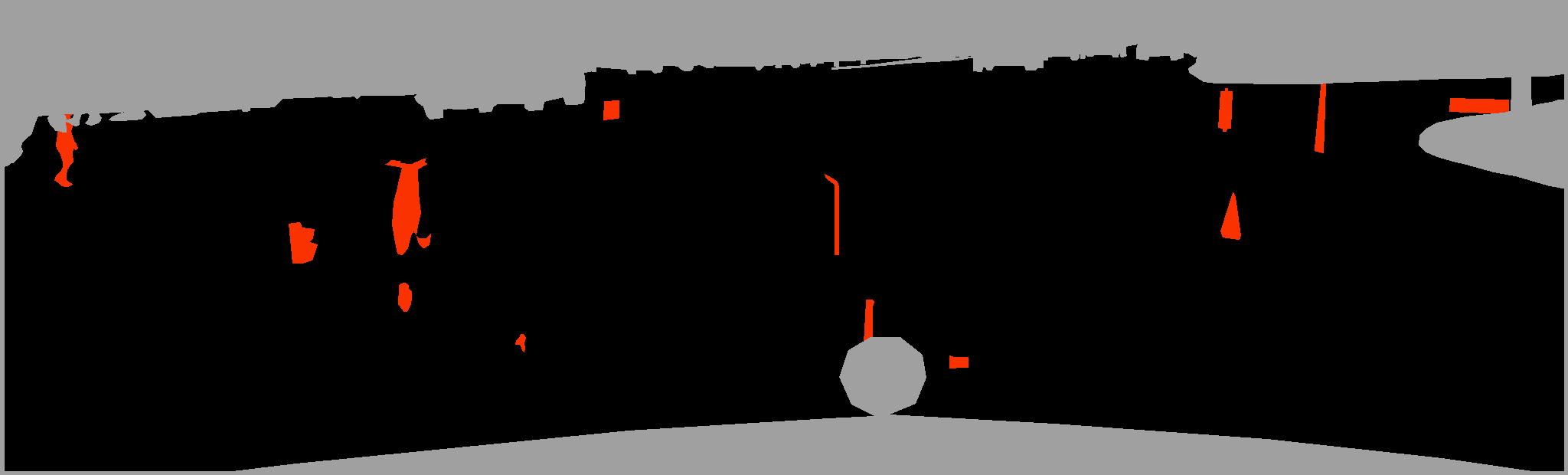}
\includegraphics[width=0.48\linewidth]{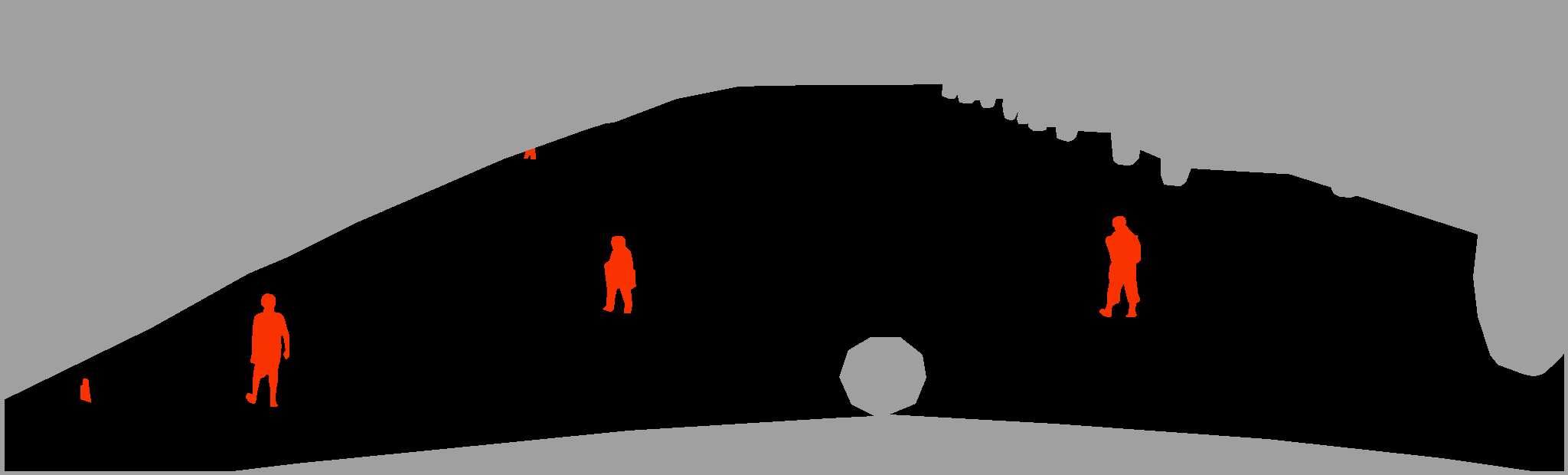}
\caption{{\bf Synthetic training obstacles.}
Using the Cityscapes dataset, we transplant random object instances onto the road to appear as small obstacles (top).
Results of the inpainting process (middle).
Labels (bottom): the discrepancy network is trained to distinguish obstacles (red) from the road area (black), while the background grey region is ignored in training.
}
\label{fig:method_train_set_synth_examples}
\end{figure}

Recall that we target unusual road obstacles that may never have been seen at training time. Therefore, we need the discrepancy network to generalize to previously-unseen objects. 

To tackle this challenge, we built a synthetic training set from only the Cityscapes~\cite{Cordts16} dataset, which contains no unusual traffic obstacles. We extracted instances of people and vehicles using the instance annotations, together with {\it traffic lights} and {\it traffic signs}, which lack instance labels, but can be extracted as connected components within their pixel-wise semantic label mask. 
Since many road obstacles are small and difficult to detect, to simulate small obstacles seen from a far distance, we selected from the whole dataset instances of size ranging from 10 to 150 pixels, and area between 100 and 5000 pix$^2$. We then sampled random objects from this database and overlaid them onto the drivable area to mimic obstacles.
Fig.~\ref{fig:method_train_set_synth_examples} features several images synthesized in this way.

\subsubsection{High frequency textures}\label{sec:method_high_freq}

The Cityscapes training set features mostly smooth road surfaces. 
Hence, test images featuring rougher road surfaces could produce false false positives. 
To prevent this, we eliminate high frequencies by means of a simple $5 \times 5$ Gaussian blur.
In the ablation study of Section~\ref{sec:exp_ablation} we show this to effective against domain shift.

\section{Experiments}
\label{sec:experiments}

In this section, we first present the  baselines, evaluation metrics, and datasets used in our experiments. We then compare our approach to the baselines and finally  perform an ablation study. 

\subsection{Baselines}
\label{sec:exp_baselines}

We compare against the following approaches. 

\parag{Resynthesis~\cite{Lis19}}
performs semantic segmentation of the image and then synthesizes an image solely from the resulting semantic map using a conditional GAN~\cite{Wang18c}. 
The differences between the original and reconstructed images are detected using
a discrepancy network similar in purpose to the one we use, but trained by synthetic anomalies;
those are produced by selecting object instances and swapping their semantic class to a random different one prior to resynthesis.

{
\parag{JSRNet \cite{Vojir21}} likewise performs image reconstruction but the bottleneck is the semantic segmenter's last layer activation.
The reconstructed image is then compared to the original one using an SSIM measure.
}

\parag{SynBoost~\cite{DiBlase21}}
expands the resynthesis approach by providing uncertainty estimates of the semantic segmentation as an additional input to its dissimilarity network,
which predicts the anomaly score.

\parag{Entropy Maximization~\cite{Chan21a}}
trains the segmentation network to maximize the output entropy on explicit OOD samples obtained by adding COCO~\cite{Lin14a} objects into Cityscapes frames. It also performs post-processing on connected components of obstacle pixels, but this part is not applicable in our per-pixel evaluation.

\parag{Learned Embedding Density ~\cite{Blum19}}
learns the inlier distribution of features extracted from a DeepLab~\cite{Chen17g} layer.
It then maps the features to latent, Gaussian-distributed vectors via a normalizing flow. The mapping is trained to maximize the likelihood of the features observed in inlier samples.

\parag{Void Classifier~\cite{Blum19}}
uses the Cityscapes {\it void} areas as examples of outliers. It can then either explicitly add {\it void} to the set of predicted classes, or learn to maximize the softmax entropy in the {\it void} regions.

\parag{Bayesian DeepLab~\cite{Mukhoti18}}
introduces dropout layers into the DeepLab network. At inference time,  it draws samples by randomizing the dropout. The uncertainty is measured as the mutual information between the resulting distribution and the network weights.

{
\parag{Maximum Softmax \cite{Hendrycks17b}}
measures the predicted softmax probability of the most probable class (max softmax) - with a low value indicating uncertain prediction and a likely semantic anomaly. 
}

\subsection{Evaluation Metrics}
\label{sec:exp_metrics}

\begin{figure} 
\centering
\includegraphics[width=0.48\linewidth]{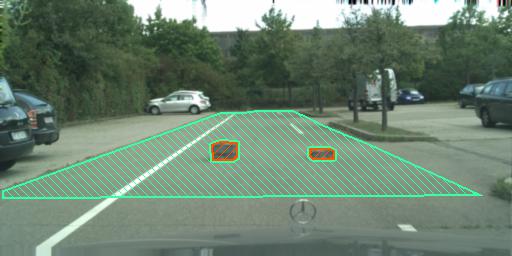}
\includegraphics[width=0.48\linewidth]{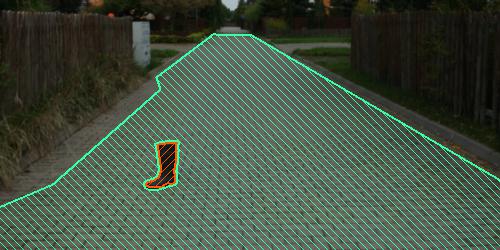}

\caption{{\bf Example ground-truth labels}.
We consider the task of distinguishing obstacle pixels (orange) from the road area (light green), while the background (unmarked) is excluded from the evaluation.
}
\label{fig:metrics_region_of_interest}
\end{figure}

We follow the evaluation protocol of the {\it Segment Me If You Can} \cite{Chan21b} benchmark and specifically the obstacle track.
Since we focus on detecting obstacles on the road, we take the Region of Interest (ROI) for evaluation purposes to be the ground-truth road area as shown in Fig.~\ref{fig:metrics_region_of_interest}. 
This restriction of ROI to drivable space follows the evaluation protocol of the original {\it Lost and Found}~\cite{Pinggera16} road obstacle benchmark and matches the area relevant to self-driving, that is, the road area where the vehicle is going to move and can encounter obstacles.
Our method, like the baselines we compare with, outputs a heatmap in the $[0,1]$ range denoting the likelihood for each pixel within the ROI of belonging to an obstacle. 
The benchmark measures performance in terms of obstacle pixel classification and obstacle instance detection.
\begin{itemize}

\item The primary pixel metrics is Average Precision (AP), that is, the area under the precision-recall curve.
This metric is more meaningful than metrics based on
the receiver operating curve (ROC) due to strong class imbalance, as obstacles typically cover less than 2\% of the total road surface.

\item A secondary pixel metric is false positive rate (FPR) at a 95\% true positive rate (TPR), which we denote as FPR$_{95}$. To compute it, the binary classification threshold is lowered until 95\% of the obstacle pixels are detected and we then measure how many false positives are introduced.

\item The main instance metric is the average $F_1$ which summarizes the numbers of true positives, false positives, and false negatives, averaged over several IoU thresholds.

\item The sIoU is the the average of the pixel intersection-over-union calculated for each obstacle instance, adjusted to account for predictions covering multiple instances, as described in~\cite{Chan21b}.

\item PPV is the positive predictive value (precision) calculated separately for each instance and then averaged.

\end{itemize}

\subsection{Datasets}
\label{sec:exp_datasets}

We report empirical results on two datasets, the well-established  {\it Fishyscapes Lost \& Found}~\cite{Pinggera16}  benchmark and a new {\it RoadObstacles21} dataset that we acquired in collaboration with colleagues at the University of Wuppertal as part of the Segment Me If You Can benchmark~\cite{Chan21b}.

\parag{Lost and Found Dataset~\cite{Pinggera16}.} 
It contains 1030 frames from image sequences captured by a vehicle approaching lost cargo items placed on parking lots and streets. We use the {\it Test No Known} subset of the test split where the training Cityscapes objects such as pedestrians and vehicles have been removed. By ensuring that all obstacles are semantic anomalies, this allows the evaluation of anomaly-detection methods in the role of obstacle detectors.

\begin{figure} 
\centering
\includegraphics[width=0.48\linewidth]{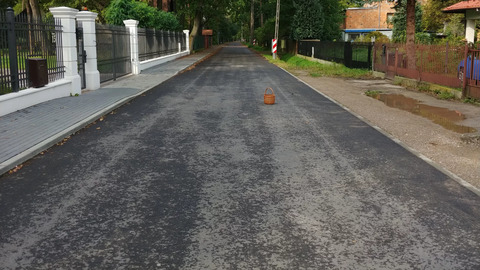}
\includegraphics[width=0.48\linewidth]{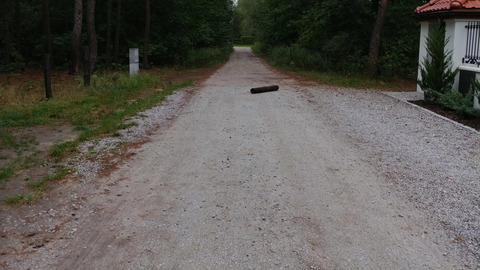}

\includegraphics[width=0.48\linewidth]{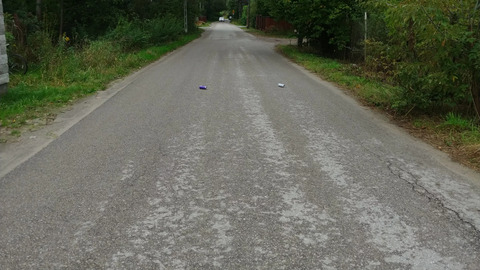}
\includegraphics[width=0.48\linewidth]{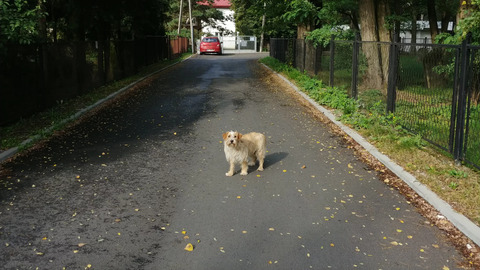}

\includegraphics[width=0.48\linewidth]{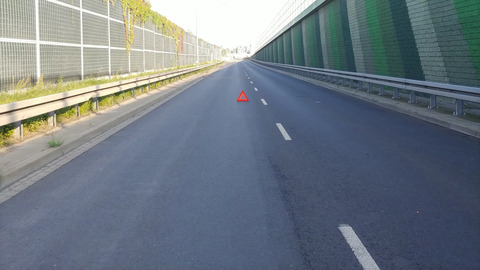}
\includegraphics[width=0.48\linewidth]{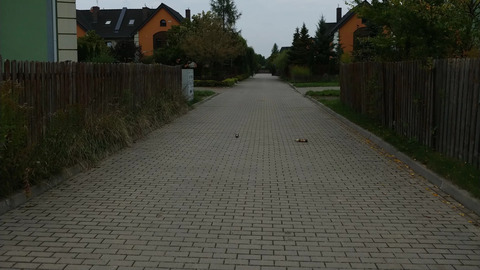}

\includegraphics[width=0.48\linewidth]{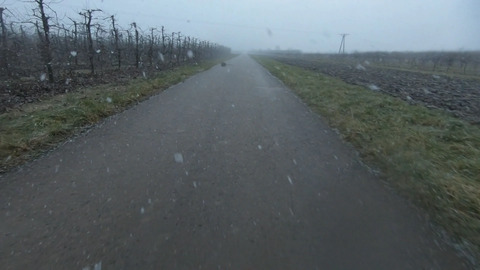}
\includegraphics[width=0.48\linewidth]{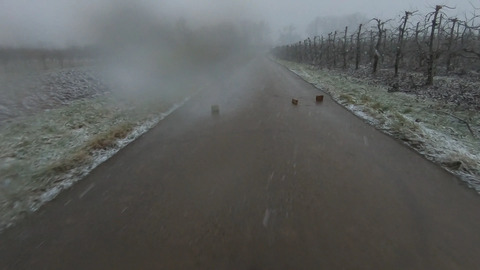}

\includegraphics[height=0.295\linewidth]{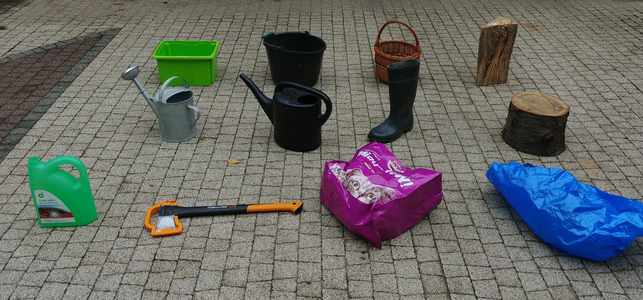}
\includegraphics[height=0.295\linewidth]{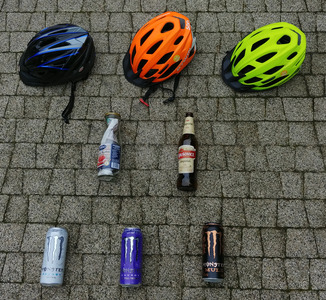}

\caption{{\bf Road Obstacles 19 dataset.}
Top: Example frames representing each of the 7 scenes.
Bottom: Some of the objects featured in the dataset.
}
\label{fig:dset_road_obstacles}
\end{figure}

\parag{RoadObstacles21 Dataset~\cite{Chan21b}.}
{\it Lost and Found} features urban environments similar to those in the {\it Cityscapes} training data.
To evaluate our obstacle detector on a wider variety of  road surfaces and objects, such as those depicted by Fig.~\ref{fig:dset_road_obstacles}, we collected our own {\it Road Obstacles} dataset. 

It features 8 different road textures and comprises a total of 327 labeled $1920 \times 1080$ frames. 
The labels include pixel masks for individual obstacle instances along with approximate outlines of the drivable area. 
We take the evaluation ROI to be the area within these outlines. If an object has a hole, such as a basket with a handle, we label the hole as outside of the ROI and ignore it in our evaluations, as we do for the background.

\subsection{Comparative Results}

\begin{table*}
\centering
\setlength{\heavyrulewidth}{2pt}
{ \small
\begin{tabular}{l|hrrrh|hrrrh}
\toprule{}   & \multicolumn{5}{c|}{Lost and Found - test no known} & \multicolumn{5}{c}{RoadObstacles21 - test} \\
{}   & \segmetricsA{c|} & \segmetricsA{c} \\
{}   & \segmetricsB & \segmetricsB \\
\midrule
Ours - road ground truth & 87.1 & 2.3 & 55.9 & 58.7 & 61.1 & 67.8 & 30.1 & 58.9 & 52.5 & 51.1 \\
\midrule
Ours - road segmentation & {\bf 82.3} & 68.5 & {\bf 49.3} & 55.6 & {\bf 54.0} & 67.5 & 32.8 & {\bf 59.3} & 52.4 & {\bf 50.8} \\
SynBoost \cite{DiBlase21} & 81.7 & {\bf 4.6} & 36.8 & {\bf 72.3} & 48.7 & 71.3 & 3.2 & 44.3 & 41.8 & 37.6 \\
Entropy Maximization \cite{Chan21a} & 77.9 & 9.7 & 45.9 & 63.1 & 49.9 & {\bf 85.1} & {\bf 0.8} & 47.9 & {\bf 62.6} & 48.5 \\
JSRNet \cite{Vojir21} & 74.2 & 6.6 & 34.3 & 45.9 & 36.0 & 28.1 & 28.9 & 18.6 & 24.5 & 11.0 \\
Embedding Density \cite{Blum19} & 61.7 & 10.4 & 37.8 & 35.2 & 27.5 & 0.8 & 46.4 & 35.6 & 2.9 & 2.3 \\
Image Resynthesis \cite{Lis19} & 57.1 & 8.8 & 27.2 & 30.7 & 19.2 & 37.7 & 4.7 & 16.6 & 20.5 & 8.4 \\
Bayesian DeepLab \cite{Mukhoti18} & 36.8 & 35.5 & 17.4 & 34.7 & 13.0 & 4.9 & 50.3 & 5.5 & 5.8 & 1.0 \\
Max Softmax \cite{Hendrycks17b} & 30.1 & 33.2 & 14.2 & 62.2 & 10.3 & 15.7 & 16.6 & 19.7 & 15.9 & 6.3 \\
Void Classifier \cite{Blum19} & 4.8 & 47.0 & 1.8 & 35.1 & 1.9 & 10.4 & 41.5 & 6.3 & 20.3 & 5.4 \\
\bottomrule
\end{tabular}
}
\vspace{1pt}
\caption{\textbf{Obstacle detection scores}. The primary metric is {\it average precision} of detecting obstacle pixels and average $F_1$ of instance detection.}
\label{table:results_segme}
\end{table*}

\begin{table*}[t]
\centering
\setlength{\heavyrulewidth}{2pt}
{ \small
\begin{tabular}{l|hrrrh|hrrrh}
\toprule{}   & \multicolumn{5}{c|}{Lost and Found - test no known} & \multicolumn{5}{c}{RoadObstacles21 - validation} \\
{}   & \segmetricsA{c|} & \segmetricsA{c} \\
{}   & \segmetricsB & \segmetricsB \\
\midrule
Ours & 82.3 & 68.5 & 49.3 & 55.6 & {\bf 54.0} & {\bf 91.0} & 100.0 & 59.4 & {\bf 57.0} & {\bf 60.0} \\
\midrule
Resynthesis & 65.0 & 100.0 & 44.2 & 35.9 & 35.6 & 83.7 & 100.0 & 55.4 & 39.6 & 45.2 \\
Blur & 78.5 & 50.7 & {\bf 49.7} & 51.0 & 50.3 & 89.3 & 100.0 & 58.4 & 51.6 & 55.3 \\
No Inpainting & 81.9 & {\bf 36.2} & 49.2 & 56.5 & 54.6 & 89.9 & 100.0 & 62.2 & 46.9 & 54.7 \\
No Discrepancy & 19.6 & 83.7 & 26.7 & 16.7 & 7.8 & 13.1 & {\bf 92.9} & {\bf 13.5} & 45.5 & 11.1 \\
Segmentation Alone & 36.5 & 100.0 & 14.4 & 91.3 & 23.1 & 28.3 & 100.0 & 22.3 & 48.8 & 26.1 \\
\midrule
No input blur & {\bf 82.4} & 100.0 & 51.5 & {\bf 56.9} & 56.2 & 87.7 & 100.0 & 63.3 & 54.3 & 59.7 \\
\bottomrule
\end{tabular}
}
\vspace{1pt}
\caption{\textbf{Ablation study results}.
\label{table:ablation_segme_val}
}
\end{table*}

We tested two versions of our method, one that operates on the road area given in the ground truth and the other on the drivable area segmented by the network of~\cite{Zhao17b}. They are referred to as ``Ours road ground truth" and ``Ours road segmentation", respectively, in \Cref{table:results_segme}.
A qualitative comparison and further examples of our method's outputs 
are shown in Fig.~\ref{fig:outputs_qualitative_comparison} and Fig.~\ref{fig:outputs_qualitative}, respectively.

\begin{figure} 
\centering
\includegraphics[width=0.48\linewidth]{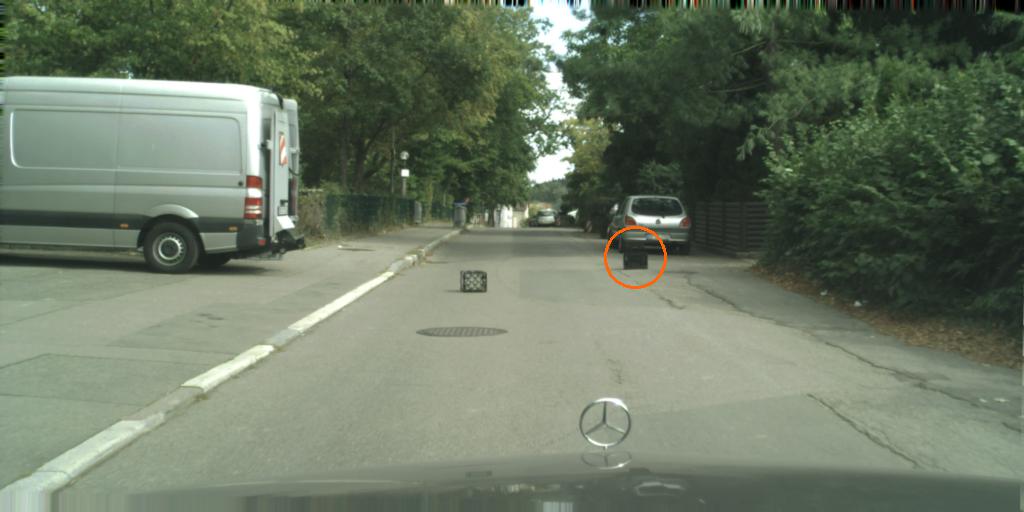}
\includegraphics[width=0.48\linewidth]{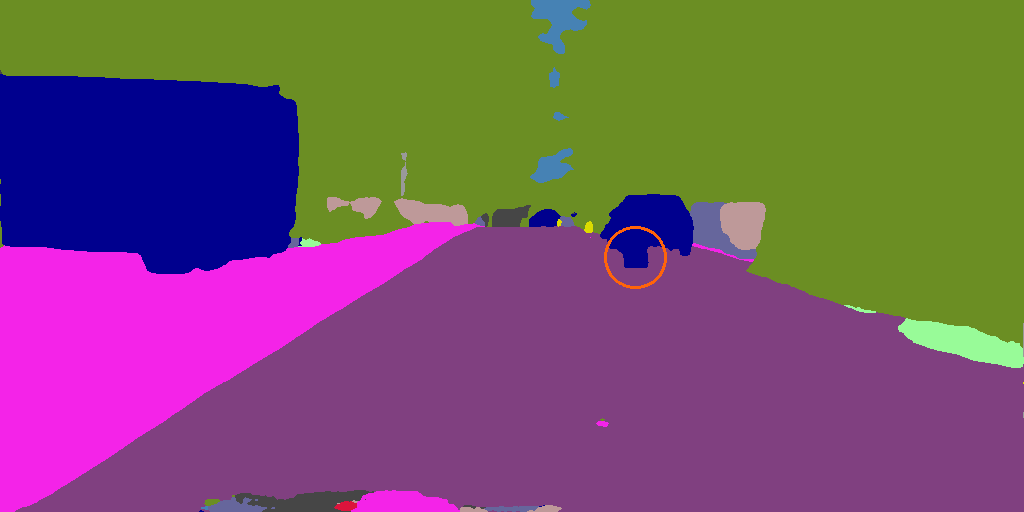}
\includegraphics[width=0.48\linewidth]{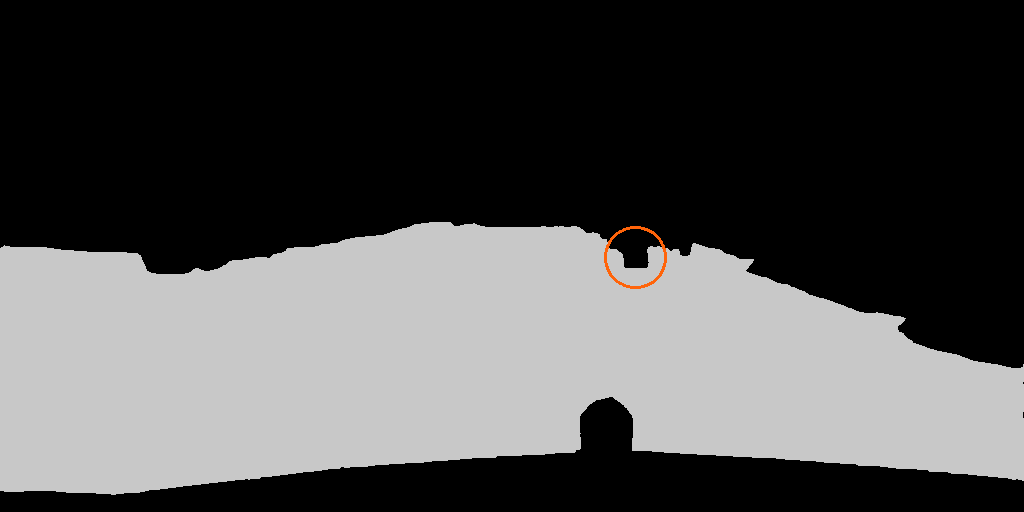}
\includegraphics[width=0.48\linewidth]{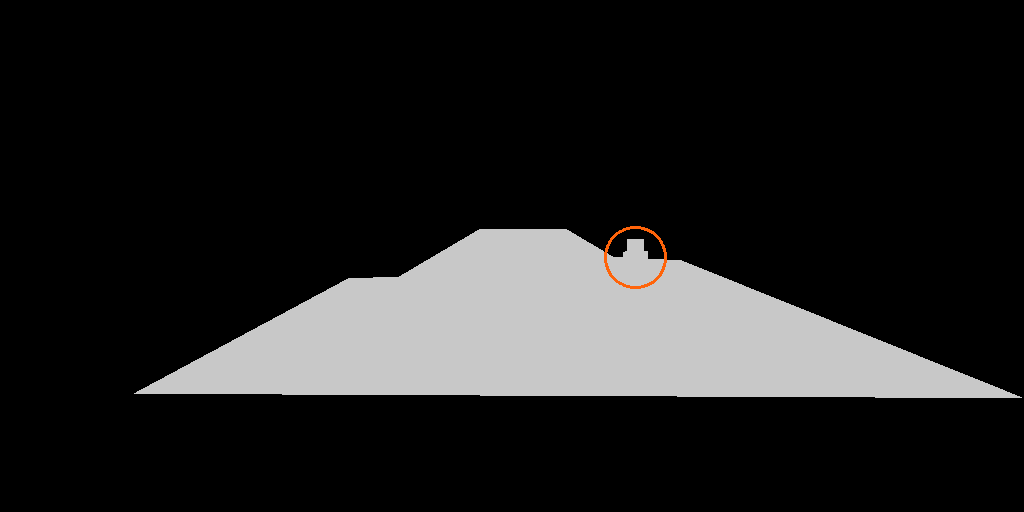}
\caption{\small{\bf Obstacle outside of predicted drivable area failure mode}. 
\textbf{Top-left:}    An obstacle is located near a vehicle shadow.
\textbf{Top-right:} The semantic segmenter classifies it as part of the background.
\textbf{Bottom-left:} Consequently it is not included in the drivable space where our obstacle detector is applied.
\textbf{Bottom-right:} Ground truth road label used as the region-of-interest for evaluation purposes. This obstacle will be counted as missed.
}
\label{fig:obstacle_outside_road}
\end{figure}

\begin{figure*} 
	\centering
	
\includegraphics[width=0.32\linewidth]{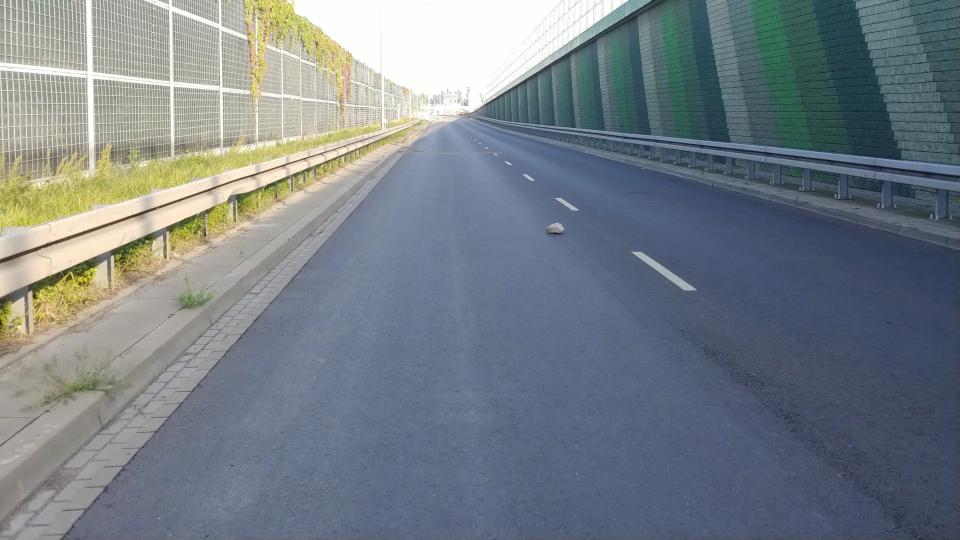}
\includegraphics[width=0.32\linewidth]{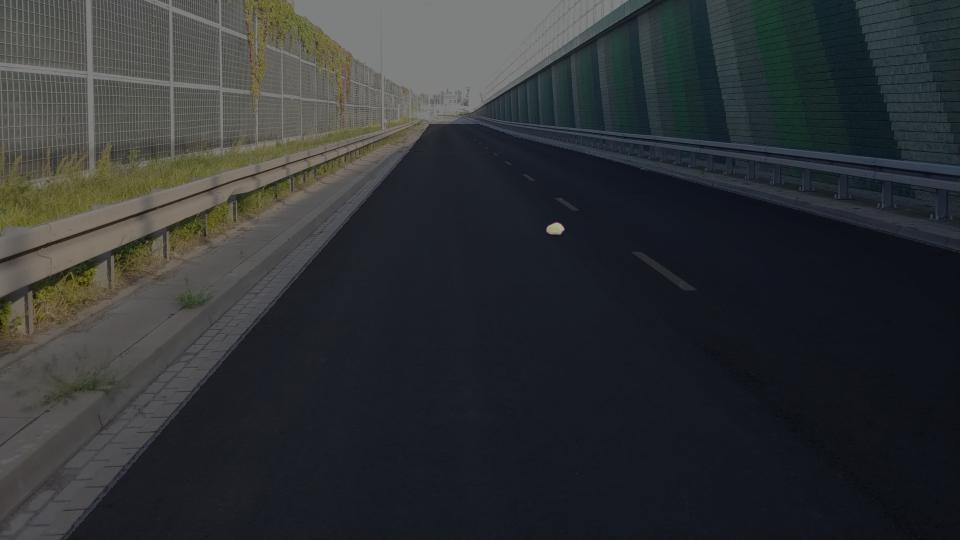}
\includegraphics[width=0.32\linewidth]{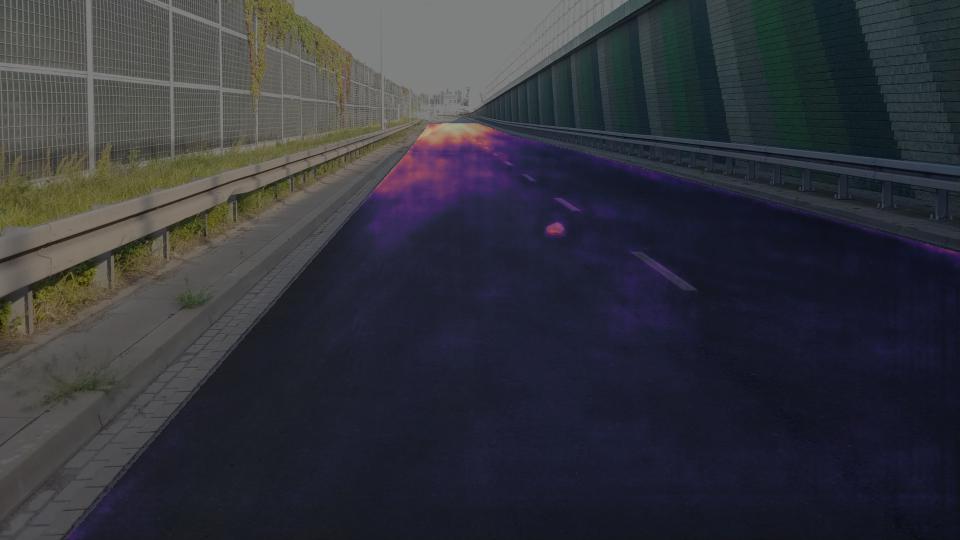}
\makebox[0.32\linewidth]{Input image}
\makebox[0.32\linewidth]{Ours}
\makebox[0.32\linewidth]{Entropy Maximization~\cite{Chan21a}}

\vspace{5pt}

\includegraphics[width=0.32\linewidth]{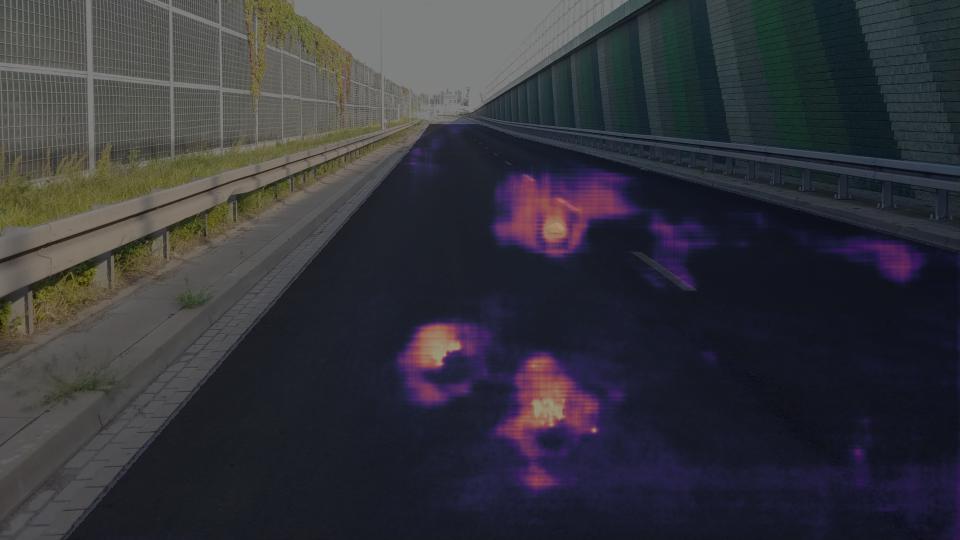}
\includegraphics[width=0.32\linewidth]{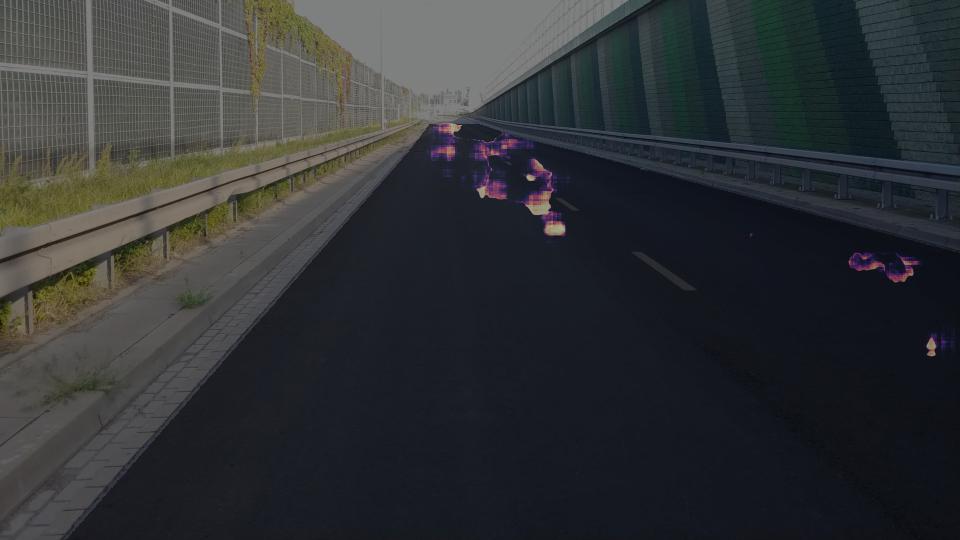}
\includegraphics[width=0.32\linewidth]{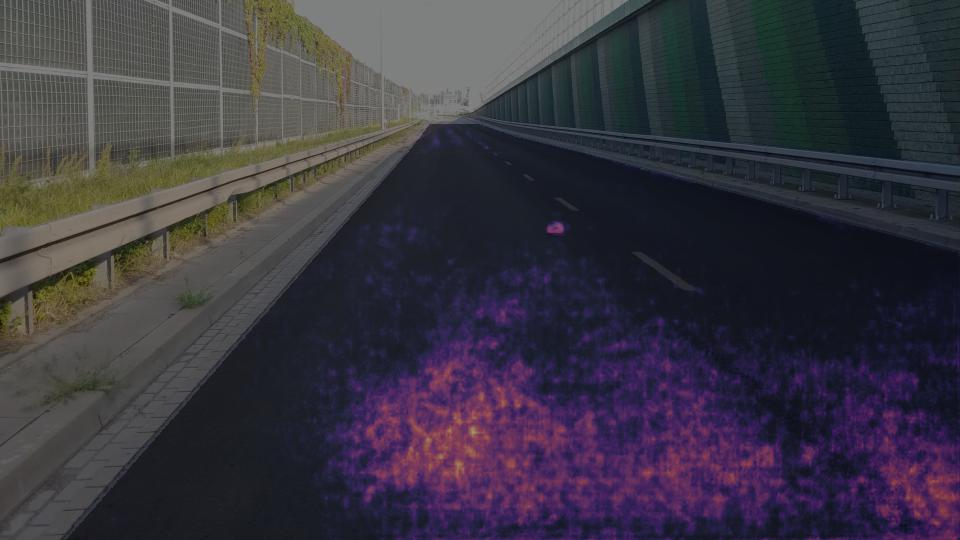}
\makebox[0.32\linewidth]{SynBoost~\cite{DiBlase21}}
\makebox[0.32\linewidth]{Resynthesis~\cite{Lis19}}
\makebox[0.32\linewidth]{Bayesian DeepLab~\cite{Mukhoti18}}

\caption{\textbf{Qualitative comparison} of method outputs. 
The darkened area corresponds to the ground-truth drivable space.
}
\label{fig:outputs_qualitative_comparison}
\end{figure*}

\begin{figure*} 
	\centering
	
\makebox[0.32\linewidth]{Input image}
\makebox[0.32\linewidth]{Inpainting of the road area}
\makebox[0.32\linewidth]{Discrepancy score}
\includegraphics[width=0.32\linewidth]{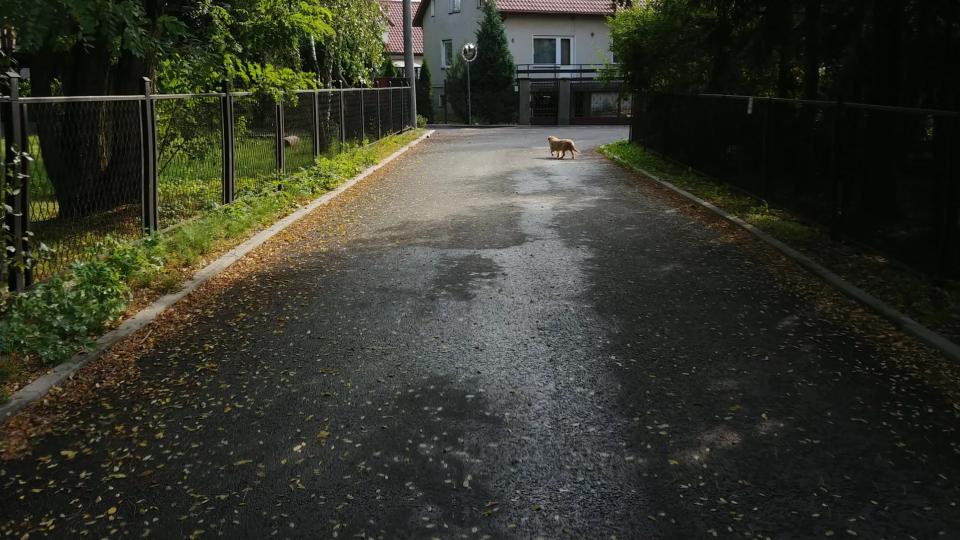}
\includegraphics[width=0.32\linewidth]{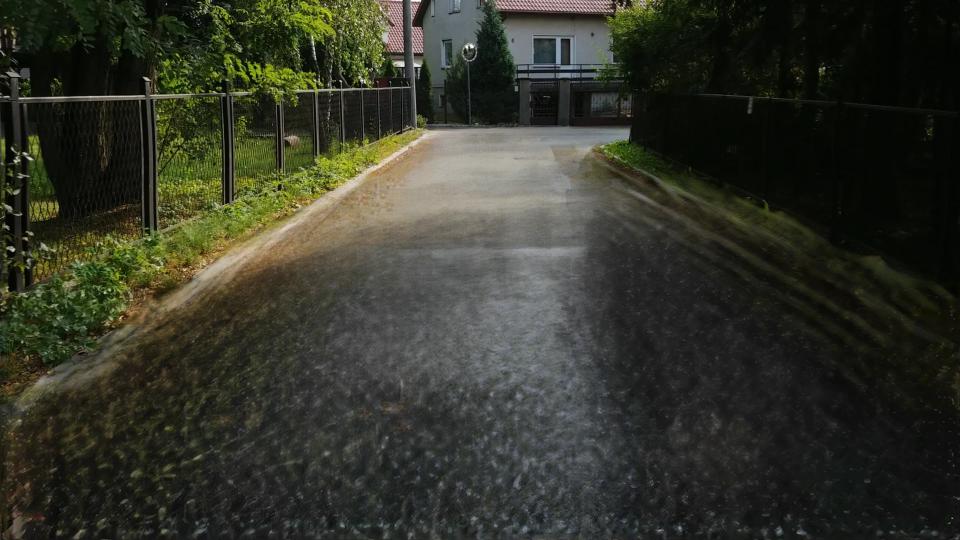}
\includegraphics[width=0.32\linewidth]{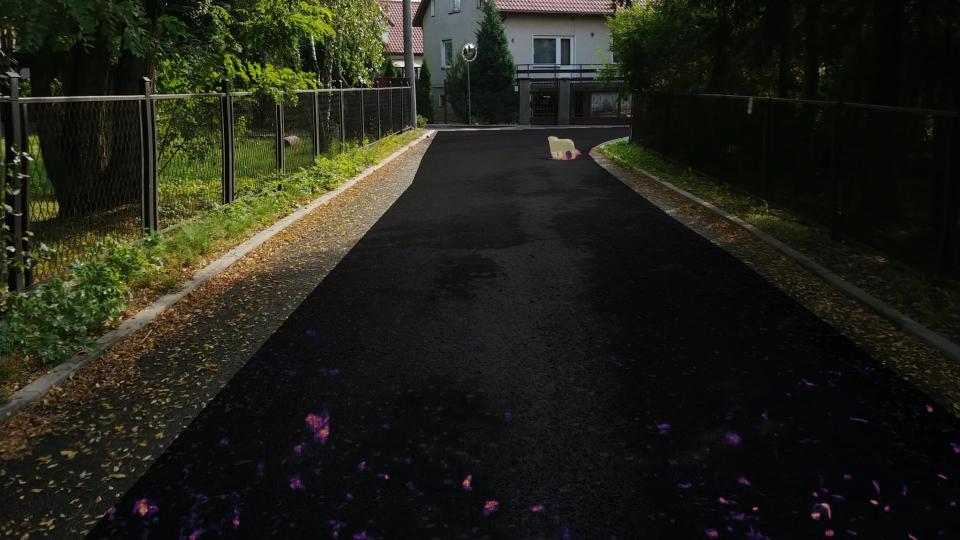}

\vspace{5pt}

\includegraphics[width=0.32\linewidth]{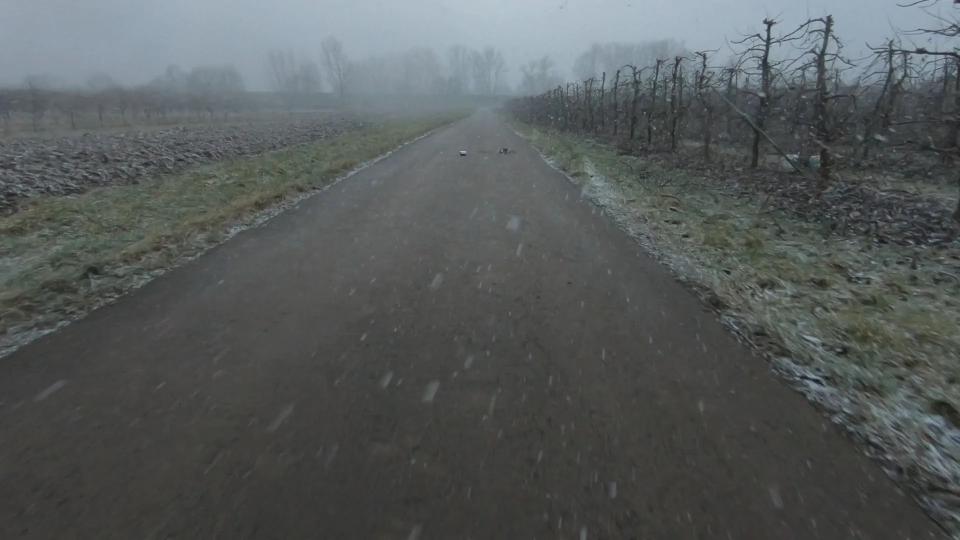}
\includegraphics[width=0.32\linewidth]{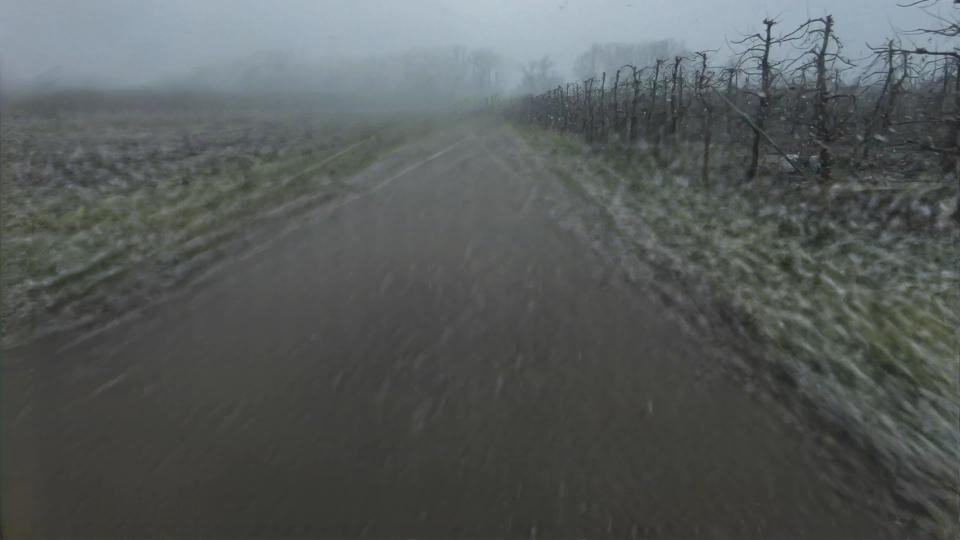}
\includegraphics[width=0.32\linewidth]{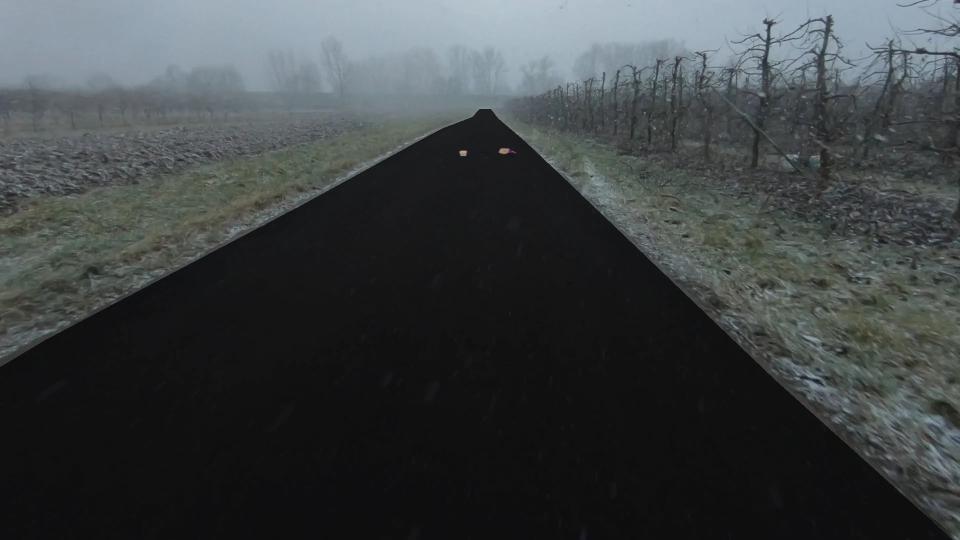}

\vspace{5pt}

\includegraphics[width=0.32\linewidth]{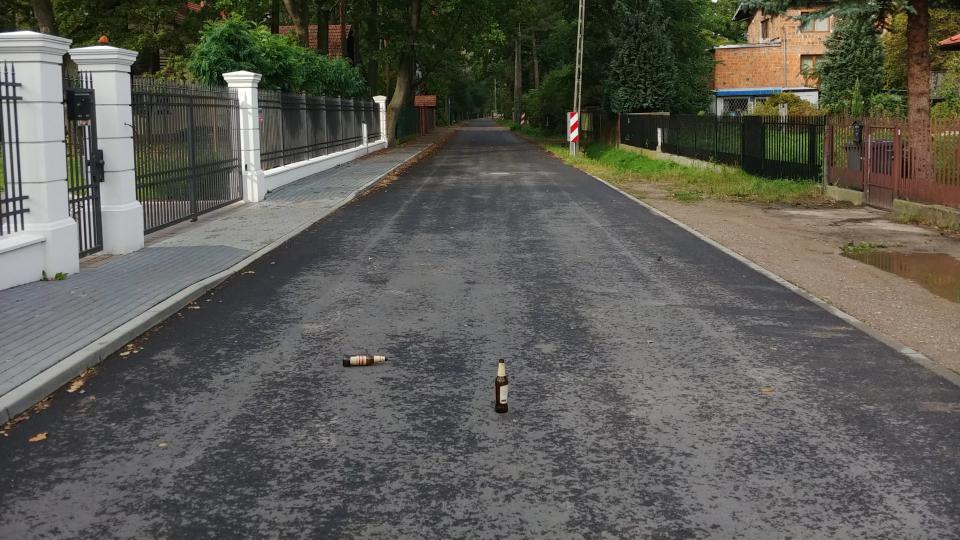}
\includegraphics[width=0.32\linewidth]{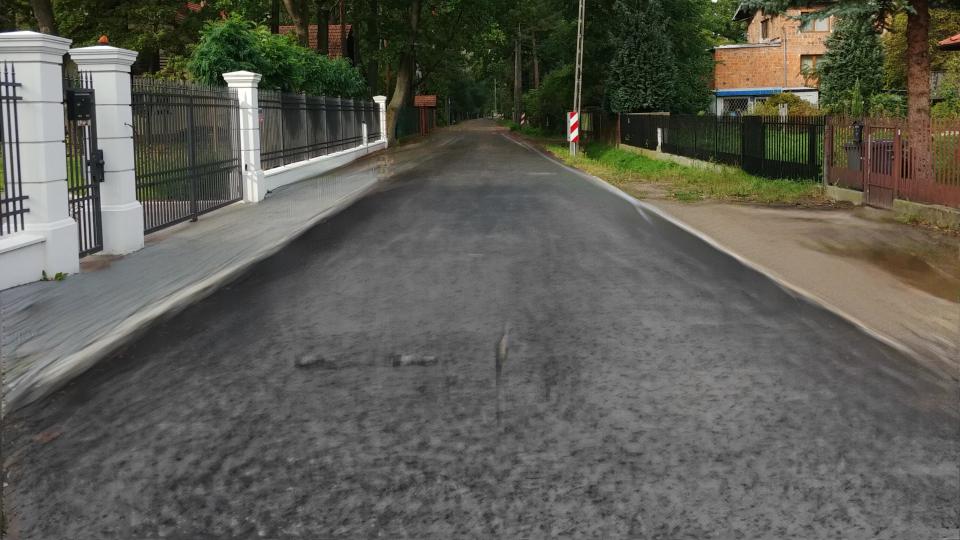}
\includegraphics[width=0.32\linewidth]{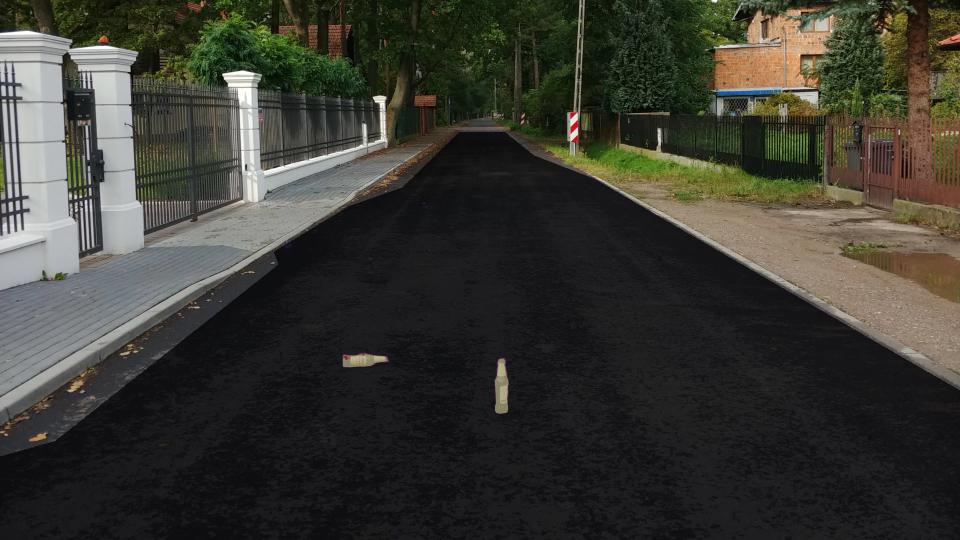}

\vspace{5pt}

\includegraphics[width=0.32\linewidth]{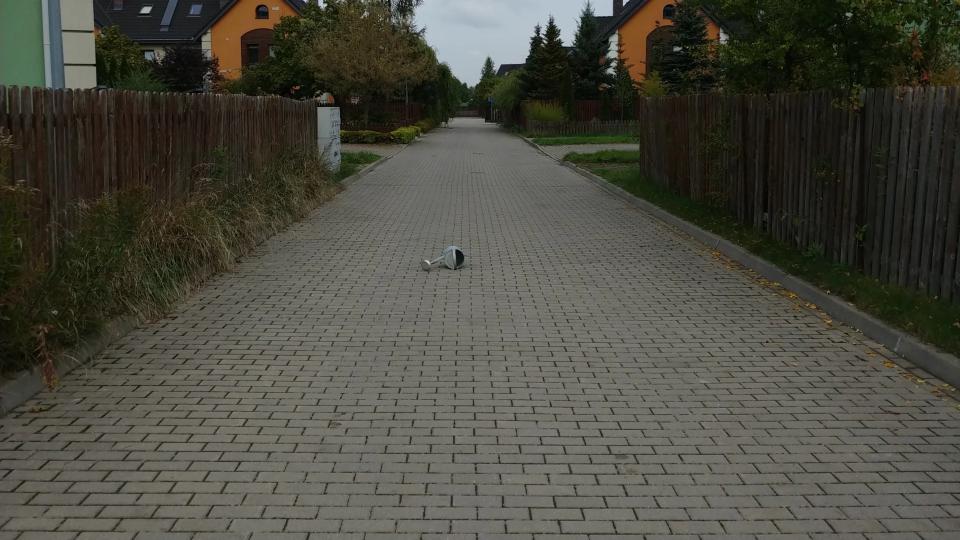}
\includegraphics[width=0.32\linewidth]{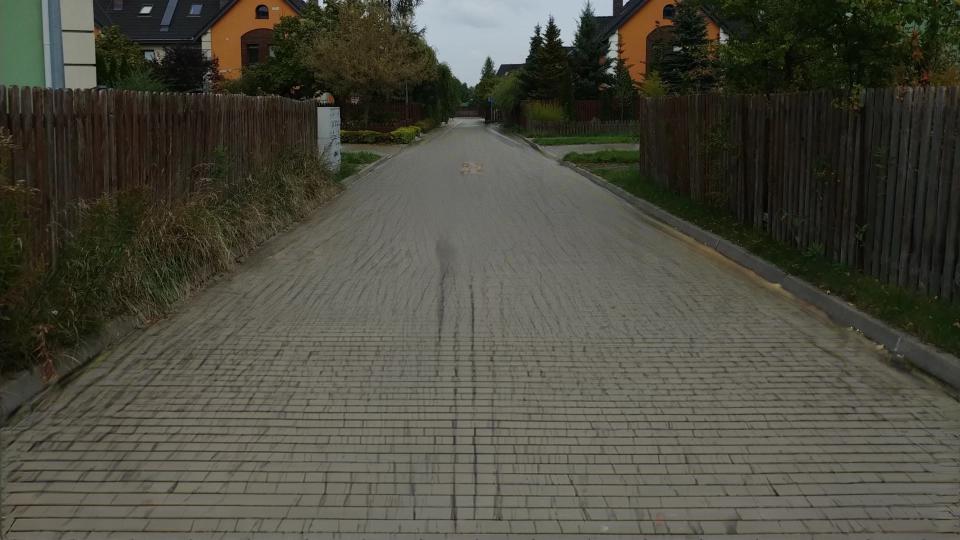}
\includegraphics[width=0.32\linewidth]{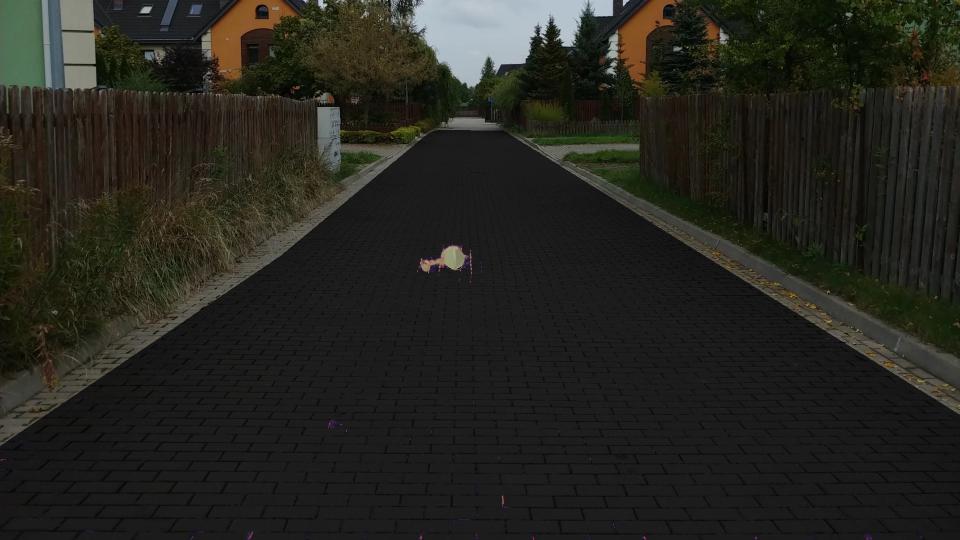}

\vspace{5pt}

\includegraphics[width=0.32\linewidth]{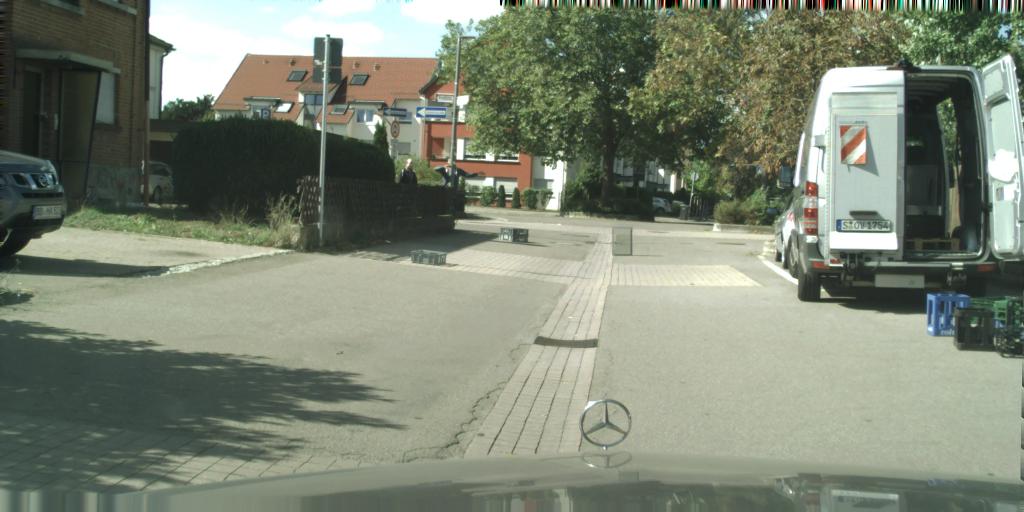}
\includegraphics[width=0.32\linewidth]{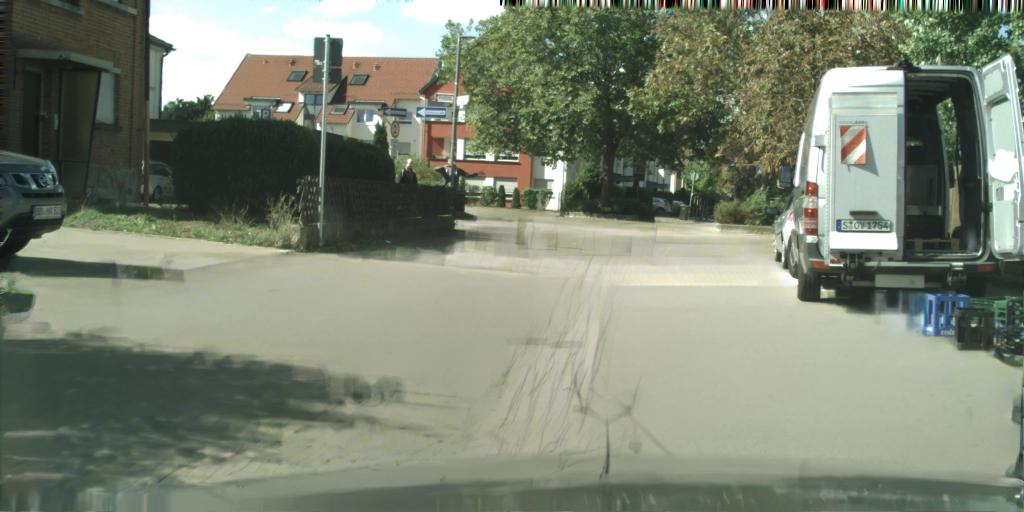}
\includegraphics[width=0.32\linewidth]{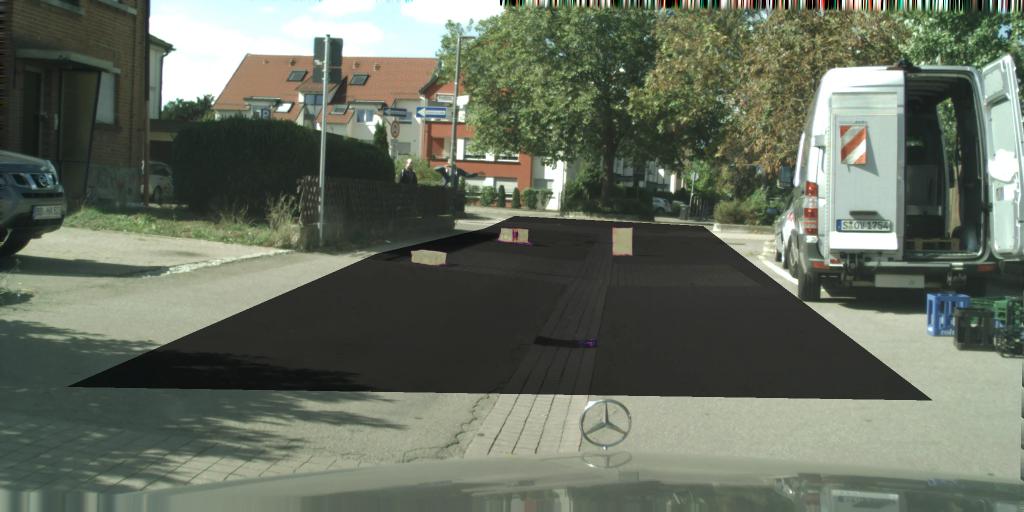}

\vspace{5pt}

\includegraphics[width=0.32\linewidth]{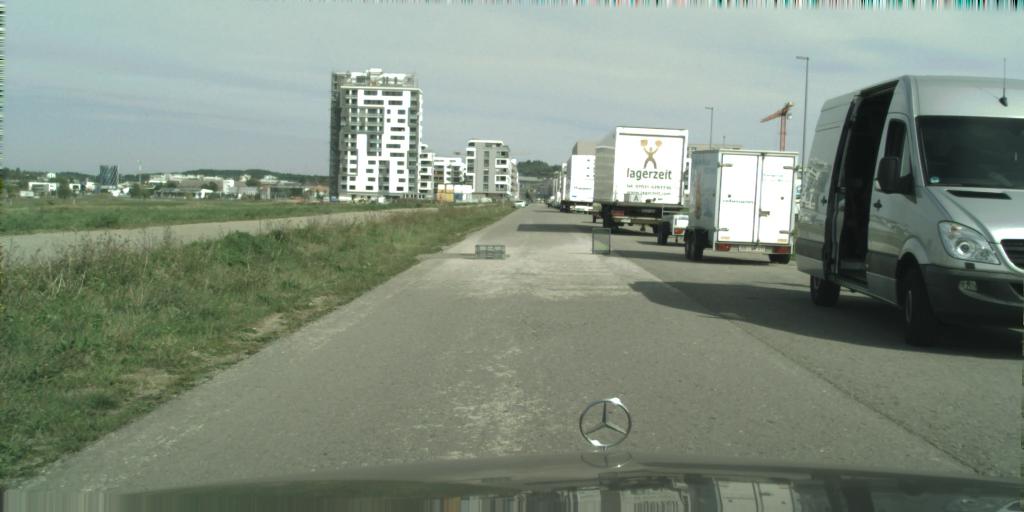}
\includegraphics[width=0.32\linewidth]{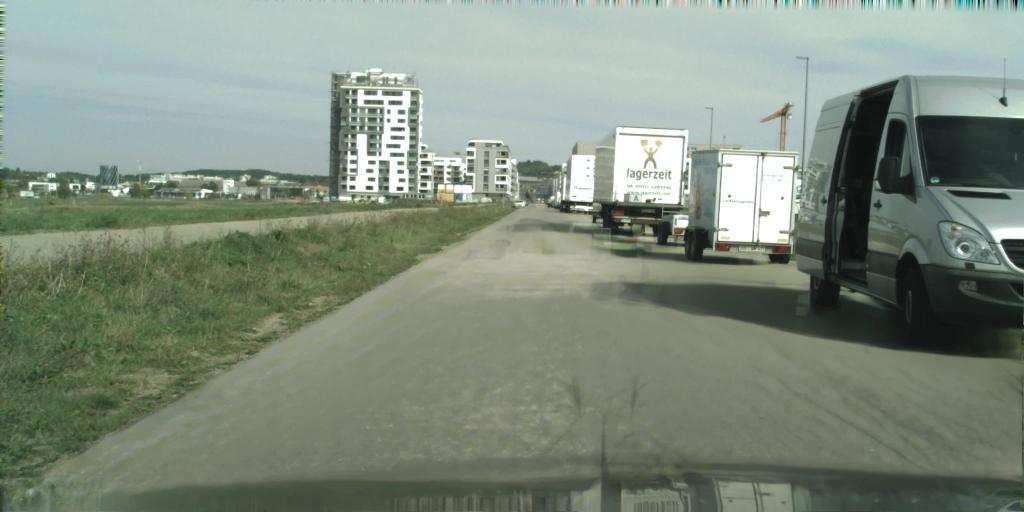}
\includegraphics[width=0.32\linewidth]{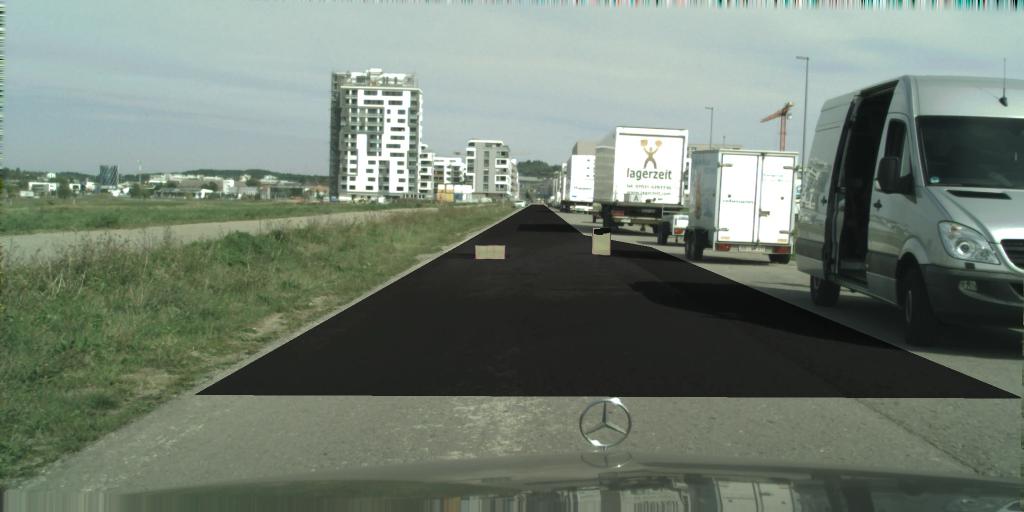}

\caption{\textbf{Example outputs of our method}. 
\textbf{Left:} Input images featuring challenging or distant obstacles.
\textbf{Center:} The result of sliding-window inpainting of the road area.
\textbf{Right:} The discrepancy score calculated by our network given the two previous images.
The darkened area corresponds to the ground-truth drivable space.
}
\label{fig:outputs_qualitative}
\end{figure*}

Our method outperform all others in terms average instance $F_1$, the primary instance metric in~\cite{Chan21b}, on both datasets.
As for pixel Average Precision (AP), it outperforms baselines on the Lost and Found dataset.
On RoadObstacles21 only two methods, SynBoost and Entropy Maximization, achieve a better AP. 
The `Ours road segmentation" variant achieves an abysmal score of the FPR$_{95}$ secondary pixel metric on Lost and Found
but the score using the ground truth is excellent. 
This points to the cause of the problem: The segmentation algorithm we use only found a part of the drivable area and, since we only look for obstacles there, we missed all those that were elsewhere, making it impossible to reach a 95\% TPR in some of the images. Fig.~\ref{fig:obstacle_outside_road} illustrates this problem. Nevertheless, even in this situation, safety is maintained because 
the vehicle controller will only drive within the predicted road space.  Hence, obstacles outside of it do not pose a risk of collision. Furthermore, as evidenced by the good results of ``Ours road ground truth", this problem will gradually fade away as road boundary detection algorithms improve.

While most methods including ours use solely the Cityscapes dataset for training, some others~\cite{Chan21a} sample the objects injected as synthetic obstacles from other datasets, such as COCO or ImageNet-1k. This strategy works well for some obstacle types, particularly the distinct and colorful ones in the {\it RoadObstacles21} set, but falls short in the {\it Fishyscapes} set. In other words, extending the original dataset with others is effective if the test obstacles are similar to the ones in the additional dataset but does not necessarily promote generalization.

\subsection{Ablation study}
\label{sec:exp_ablation}

In \Cref{table:ablation_segme_val}, we report  the results of an ablation study  during which the discrepancy network was retrained with selected components altered or disabled. 

In the {\it Resynthesis} variant, the inpainting described in Section~\ref{sec:method_inpainting} is replaced by an image synthesizer~\cite{Wang18c} from predicted semantic labels as in~\cite{Lis19}. While the inpainter can reconstruct novel road textures based on the visual context, the generator produces a texture similar to the training roads, and this is reflected by degraded performance on {\it Road Obstacles}.

The {\it Blur} variant replaces inpainting by a Gaussian blur operation. We took the kernel size to be 200 to match the size of the inpainting window.
The {\it No Inpainting} variant does away with inpainting. We keep the architecture unchanged, but pass two copies of the image into both input streams and retrain the network. This also degrades performance, thus confirming the importance of inpainting. We can also do the reverse, that is, remove the discrepancy network, and compute the $L_1$ distance between the RGB values of the input image and the inpainted result. The results of the corresponding {\it No Discrepancy} variant are even worse. 

The \textit{Segmentation Alone} entry corresponds to detections made by finding groups of non-road pixels enclosed within the road region detected by the segmentation algorithm~\cite{Zhao17b}. This by itself clearly fails, thereby justifying the extra step we propose in this paper.

\subsection{Limitations}

\textbf{Erasing Safe Regions.}
The out-of-distribution regions detected by our approach can include both genuine obstacles and distinctive but safe road regions such as manholes, road cracks, and leaves. 
The discrepancy network of \Cref{sec:method_discrepancy} alleviates this problem by learning what altered regions can be safely ignored, but that relies on examples being present in the training set.
For further improvements, an approach would be to look into methods that expand the training set, such as ~\cite{Grcic21,Grcic21b}.

\textbf{Perspective effects.}
Our approach does not explicitly account for obstacles that are a further to be smaller, due to perspective effects. Correcting for this, as we did in~\cite{Lis23a}, should also help reduce false-positives, especially in regions close to the camera. 

\textbf{Computational Cost.}
The sliding window inpainting algorithm is computationally expensive, requiring multiple executions of the inpainter.
The current naive implementation takes 6.7 s (inpainting) plus 0.2 s (discrepancy network) per frame on a single A100 GPU, as measured on the {\it RoadObstacles21} validation dataset.
As this work is intended as a proof-of-concept,  we used off-the-shelf inpainting code without production optimization. One way to remedy this would be to switch to 16-bites precision, which is much more efficient on modern GPUs. Another would be to parallelize the inpainting.
Furthermore, even the slow version of our approach can be used effectively: 
Since our algorithm detects previously unseen obstacles,  it can be used to extract obstacles from unlabeled videos for additional supervised training, even without real-time inference.

\section{Conclusion}\label{sec:conclusion}

We have introduced a pipeline capable of detecting road obstacles in driving scenarios, given only single monocular images as input.
We perform inpainting of the drivable area, erasing the localized obstacles while preserving the road surface.
Our discrepancy network learns to accurately detect the removed obstacles and ignore irrelevant artifacts of reconstruction.
This detector, trained only with synthetically altered Cityscapes data, is capable of generalizing to a variety of real-world obstacles and road surfaces.
We have demonstrated this on the {\it Fishyscapes Lost \& Found} benchmark, as well as on our newly collected dataset.
We will release our dataset to the community to help researchers in this field.

In the future, we fill work on coupling more tightly the algorithms that detects the drivable area and ours for improved results and for potential integration into a full self-driving system. We also plan to exploit adaptive inpainting patch sizes to account for perspective distortion.

\subsection*{Acknowledgements}

We thank {\bf Hermann Blum} for adapting the Fishyscapes benchmark and baseline methods to our data as well as help with the evaluations.
We thank {\bf Mirosław Lis} for assistance in the collection of the {\it Road Obstacles} dataset.

\clearpage

{
\small
\bibliographystyle{IEEEtran}
\bibliography{bib/string,bib/vision,bib/learning,bib/misc,bib/vehicle_industry}

\begin{thebibliography}{10}
\providecommand{\url}[1]{#1}
\csname url@samestyle\endcsname
\providecommand{\newblock}{\relax}
\providecommand{\bibinfo}[2]{#2}
\providecommand{\BIBentrySTDinterwordspacing}{\spaceskip=0pt\relax}
\providecommand{\BIBentryALTinterwordstretchfactor}{4}
\providecommand{\BIBentryALTinterwordspacing}{\spaceskip=\fontdimen2\font plus
\BIBentryALTinterwordstretchfactor\fontdimen3\font minus
  \fontdimen4\font\relax}
\providecommand{\BIBforeignlanguage}[2]{{%
\expandafter\ifx\csname l@#1\endcsname\relax
\typeout{** WARNING: IEEEtran.bst: No hyphenation pattern has been}%
\typeout{** loaded for the language `#1'. Using the pattern for}%
\typeout{** the default language instead.}%
\else
\language=\csname l@#1\endcsname
\fi
#2}}
\providecommand{\BIBdecl}{\relax}
\BIBdecl

\bibitem{Chen17rbnet}
Z.~Chen and Z.~Chen, ``{RBNet: A Deep Neural Network for Unified Road and Road
  Boundary Detection},'' in \emph{International Conference on Neural
  Information Processing}, 2017.

\bibitem{Munoz17deep}
J.~Munoz-Bulnes, C.~Fernandez, I.~Parra, D.~Fern{\'a}ndez-Llorca, and M.~A.
  Sotelo, ``{Deep fully convolutional networks with random data augmentation
  for enhanced generalization in road detection},'' in \emph{IEEE International
  Conference on Intelligent Transportation Systems (ITSC)}, 2017.

\bibitem{Liu21end}
R.~Liu, Z.~Yuan, T.~Liu, and Z.~Xiong, ``{End-to-end lane shape prediction with
  transformers},'' in \emph{IEEE Winter Conference on Applications of Computer
  Vision}, 2021, pp. 3694--3702.

\bibitem{Qin20a}
Z.~Qin, H.~Wang, and X.~Li, ``{Ultra Fast Structure-Aware Deep Lane
  Detection},'' in \emph{European Conference on Computer Vision}, 2020.

\bibitem{Ren15}
S.~Ren, K.~He, R.~Girshick, and J.~Sun, ``{Faster {R-CNN}: Towards Real-Time
  Object Detection with Region Proposal Networks},'' in \emph{Advances in
  Neural Information Processing Systems}, 2015.

\bibitem{Dollar12b}
P.~Doll{\'a}r, C.~Wojek, B.~Schiele, and P.~Perona, ``{Pedestrian Detection: An
  Evaluation of the State of the Art},'' \emph{IEEE Transactions on Pattern
  Analysis and Machine Intelligence}, vol.~34, no.~4, pp. 743--761, 2012.

\bibitem{Chabot17manta}
F.~Chabot, M.~Chaouch, J.~Rabarisoa, C.~Teuliere, and T.~Chateau, ``Deep manta:
  A coarse-to-fine many-task network for joint 2d and 3d vehicle analysis from
  monocular image,'' in \emph{Conference on Computer Vision and Pattern
  Recognition}, 2017.

\bibitem{Zhao17b}
H.~Zhao, J.~Shi, X.~Qi, X.~Wang, and J.~Jia, ``{Pyramid Scene Parsing
  Network},'' in \emph{Conference on Computer Vision and Pattern Recognition},
  2017.

\bibitem{Chen17a}
L.-C. Chen, G.~Papandreou, F.~Schroff, and H.~Adam, ``{Rethinking Atrous
  Convolution for Semantic Image Segmentation},'' in \emph{arXiv Preprint},
  2017.

\bibitem{Romera17}
E.~Romera, J.~M. Alvarez, L.~M. Bergasa, and R.~Arroyo, ``{ERFNet: Efficient
  Residual Factorized Convnet for Real-Time Semantic Segmentation},''
  \emph{IEEE Transactions on Intelligent Transportation Systems}, 2017.

\bibitem{He17a}
K.~He, G.~Gkioxari, P.~Doll{\'a}r, and R.~B. Girshick, ``{Mask R-CNN},'' in
  \emph{International Conference on Computer Vision}, 2017.

\bibitem{Kirillov19}
A.~Kirillov, K.~He, R.~Girshick, C.~Rother, and P.~Doll{\'a}r, ``{Panoptic
  Segmentation},'' in \emph{Conference on Computer Vision and Pattern
  Recognition}, 2019.

\bibitem{Xiong19b}
Y.~Xiong, R.~Liao, H.~Zhao, R.~Hu, M.~Bai, E.~Yumer, and R.~Urtasun, ``{UPSNet:
  A Unified Panoptic Segmentation Network},'' in \emph{Conference on Computer
  Vision and Pattern Recognition}, 2019.

\bibitem{LaneAssist}
``{AutoNation Drive Editors' Guide to Lane Assist Systems},''
  \url{https://www.autonationdrive.com/research/best-cars-with-lane-assist.htm},
  2021.

\bibitem{SinghKamireddypalli20}
A.~Singh, A.~Kamireddypalli, V.~Gandhi, and K.~M. Krishna, ``{Lidar guided
  small obstacle segmentation},'' in \emph{International Conference on
  Intelligent Robots and Systems}.\hskip 1em plus 0.5em minus 0.4em\relax IEEE,
  2020, pp. 8513--8520.

\bibitem{Pinggera16}
P.~Pinggera, S.~Ramos, S.~Gehrig, U.~Franke, C.~Rother, and R.~Mester, ``{Lost
  and Found: Detecting Small Road Hazards for Self-Driving Vehicles},'' in
  \emph{International Conference on Intelligent Robots and Systems}, 2016.

\bibitem{Zavrtanik20}
V.~Zavrtanik, M.~Kristan, and D.~Skoaj, ``{Reconstruction by Inpainting for
  Visual Anomaly Detection},'' \emph{Pattern Recognition}, 2020.

\bibitem{Haselmann18}
M.~Haselmann, D.~P. Gruber, and P.~Tabatabai, ``{Anomaly Detection Using Deep
  Learning Based Image Completion},'' in \emph{International Conference on
  Machine Learning}, 2018.

\bibitem{Bhattad18}
A.~Bhattad, J.~Rock, and D.~Forsyth, ``{Detecting Anomalous Faces with 'No
  Peeking' Autoencoders},'' in \emph{Conference on Computer Vision and Pattern
  Recognition}, 2018.

\bibitem{Cordts16}
M.~Cordts, M.~Omran, S.~Ramos, T.~Rehfeld, M.~Enzweiler, R.~Benenson,
  U.~Franke, S.~Roth, and B.~Schiele, ``{The Cityscapes Dataset for Semantic
  Urban Scene Understanding},'' in \emph{Conference on Computer Vision and
  Pattern Recognition}, 2016.

\bibitem{Lis19}
K.~Lis, K.~Nakka, M.~Salzmann, and P.~Fua, ``{Detecting the Unexpected via
  Image Resynthesis},'' in \emph{International Conference on Computer Vision},
  2019.

\bibitem{Blum19}
H.~Blum, P.-E. Sarlin, J.~Nieto, R.~Siegwart, and C.~Cadena, ``{Fishyscapes: A
  Benchmark for Safe Semantic Segmentation in Autonomous Driving},'' in
  \emph{International Conference on Computer Vision}, October 2019.

\bibitem{Malinin18}
A.~Malinin and M.~Gales, ``{Predictive Uncertainty Estimation via Prior
  Networks},'' in \emph{Advances in Neural Information Processing Systems},
  2018.

\bibitem{Bevandic19}
P.~Bevandi{\'c}, I.~Kreso, M.~Orsic, and S.~Segvi{\'c}, ``{Simultaneous
  Semantic Segmentation and Outlier Detection in Presence of Domain Shift},''
  in \emph{German Conference on Pattern Recognition}, 2019.

\bibitem{Ramos17}
S.~Ramos, S.~Gehrig, P.~Pinggera, U.~Franke, and C.~Rother, ``{Detecting
  Unexpected Obstacles for Self-Driving Cars: Fusing Deep Learning and
  Geometric Modeling},'' in \emph{IEEE Intelligent Vehicles Symposium}, 2017.

\bibitem{Gupta18}
A.~Gupta, J.~Johnson, L.~Fei-Fei, S.~Savarese, and A.~Alahi, ``{Social GAN:
  Socially Acceptable Trajectories with Generative Adversarial Networks},'' in
  \emph{Conference on Computer Vision and Pattern Recognition}, 2018.

\bibitem{Xue19}
F.~Xue, A.~Ming, M.~Zhou, and Y.~Zhou, ``{A Novel Multi-Layer Framework for
  Tiny Obstacle Discovery},'' in \emph{International Conference on Robotics and
  Automation}, 2019.

\bibitem{Akcay18}
S.~Akcay, A.~Atapour-Abarghouei, and T.~P. Breckon, ``{Ganomaly:
  Semi-Supervised Anomaly Detection via Adversarial Training},'' in \emph{Asian
  Conference on Computer Vision}, 2018.

\bibitem{Creusot15}
C.~Creusot and A.~Munawar, ``{Real-Time Small Obstacle Detection on Highways
  Using Compressive RBM Road Reconstruction},'' in \emph{Intelligent Vehicles
  Symposium}, 2015.

\bibitem{Munawar17}
A.~Munawar, P.~Vinayavekhin, and G.~{De Magistris}, ``{Limiting the
  Reconstruction Capability of Generative Neural Network Using Negative
  Learning},'' in \emph{IEEE International Workshop on Machine Learning for
  Signal Processing}, 2017.

\bibitem{Schlegl17}
T.~Schlegl, P.~Seeb{\"o}ck, S.~M. Waldstein, U.~Schmidt-Erfurth, and G.~Langs,
  ``{Unsupervised Anomaly Detection with Generative Adversarial Networks to
  Guide Marker Discovery},'' in \emph{International Conference on Information
  Processing in Medical Imaging}, 2017.

\bibitem{Xia20}
Y.~Xia, Y.~Zhang, F.~Liu, W.~Shen, and A.~Yuille, ``{Synthesize Then Compare:
  Detecting Failures and Anomalies for Semantic Segmentation},'' in
  \emph{European Conference on Computer Vision}, 2020.

\bibitem{Isola16}
P.~Isola, J.~Zhu, T.~Zhou, and A.~Efros, ``{Image-To-Image Translation with
  Conditional Adversarial Networks},'' in \emph{arXiv Preprint}, 2016.

\bibitem{Wang18c}
T.-C. Wang, M.-Y. Liu, J.-Y. Zhu, A.~Tao, J.~Kautz, and B.~Catanzaro,
  ``{High-Resolution Image Synthesis and Semantic Manipulation with Conditional
  {GANs}},'' in \emph{Conference on Computer Vision and Pattern Recognition},
  2018.

\bibitem{Vojir21}
T.~Vojir, T.~{\v S}ipka, R.~Aljundi, N.~Chumerin, D.~O. Reino, and J.~Matas,
  ``{Road Anomaly Detection by Partial Image Reconstruction with Segmentation
  Coupling},'' in \emph{International Conference on Computer Vision}, October
  2021.

\bibitem{Munawar15}
A.~Munawar and C.~Creusot, ``{Structural Inpainting of Road Patches for Anomaly
  Detection},'' in \emph{IAPR International Conference on Machine Vision
  Applications}, 2015.

\bibitem{Bescos18}
B.~Bescos, J.~Neira, R.~Siegwart, and C.~Cadena, ``{Empty Cities: Image
  Inpainting for a Dynamic-Object-Invariant Space},'' in \emph{International
  Conference on Robotics and Automation}, 2019.

\bibitem{Berlincioni19}
L.~Berlincioni, F.~Becattini, L.~Galteri, L.~Seidenari, and A.~{Del Bimbo},
  ``{Road Layout Understanding by Generative Adversarial Inpainting},'' in
  \emph{Inpainting and Denoising Challenges}, 2019, pp. 111--128.

\bibitem{Deng09}
J.~Deng, W.~Dong, R.~Socher, L.-J. Li, K.~Li, and L.~Fei-Fei, ``{Imagenet: A
  Large-Scale Hierarchical Image Database},'' in \emph{Conference on Computer
  Vision and Pattern Recognition}, 2009.

\bibitem{Neuhold17}
G.~Neuhold, T.~Ollmann, S.~{Rota Bulo}, and P.~Kontschieder, ``{The Mapillary
  Vistas Dataset for Semantic Understanding of Street Scenes},'' in
  \emph{Conference on Computer Vision and Pattern Recognition}, 2017.

\bibitem{Srivastava14}
N.~Srivastava, G.~Hinton, A.~Krizhevsky, I.~Sutskever, and R.~Salakhutdinov,
  ``{Dropout: A Simple Way to Prevent Neural Networks from Overfitting},''
  \emph{Journal of Machine Learning Research}, vol.~15, pp. 1929--1958, 2014.

\bibitem{Kendall15b}
A.~Kendall, V.~Badrinarayanan, and R.~Cipolla, ``{Bayesian Segnet: Model
  Uncertainty in Deep Convolutional Encoder-Decoder Architectures for Scene
  Understanding},'' in \emph{arXiv Preprint}, 2015.

\bibitem{Kendall17}
A.~Kendall and Y.~Gal, ``{What Uncertainties Do We Need in Bayesian Deep
  Learning for Computer Vision?}'' in \emph{Advances in Neural Information
  Processing Systems}, 2017.

\bibitem{Lakshminarayanan17}
B.~Lakshminarayanan, A.~Pritzel, and C.~Blundell, ``{Simple and Scalable
  Predictive Uncertainty Estimation Using Deep Ensembles},'' in \emph{Advances
  in Neural Information Processing Systems}, 2017.

\bibitem{Gustafsson20}
F.~K. Gustafsson, M.~Danelljan, and T.~B. Sch{\"o}n, ``{Evaluating Scalable
  Bayesian Deep Learning Methods for Robust Computer Vision},'' in
  \emph{Conference on Computer Vision and Pattern Recognition}, 2020.

\bibitem{Chan21a}
R.~Chan, M.~Rottmann, and H.~Gottschalk, ``{Entropy Maximization and Meta
  Classification for Out-Of-Distribution Detection in Semantic Segmentation},''
  in \emph{International Conference on Computer Vision}, 2021.

\bibitem{Lin14a}
T.-Y. Lin, M.~Maire, S.~Belongie, J.~Hays, P.~Perona, D.~Ramanan,
  P.~Doll{\'a}r, and C.~Zitnick, ``{Microsoft COCO: Common Objects in
  Context},'' in \emph{European Conference on Computer Vision}, 2014, pp.
  740--755.

\bibitem{Grcic21}
M.~Grcic, P.~Bevandic, and S.~Segvic, ``{Dense Open-set Recognition with
  Synthetic Outliers Generated by Real NVP},'' in \emph{VISIGRAPP (4: VISAPP)},
  2021.

\bibitem{Grcic21b}
\BIBentryALTinterwordspacing
M.~Grcic, P.~Bevandic, Z.~Kalafatic, and S.~Segvic, ``{Dense anomaly detection
  by robust learning on synthetic negative data},'' in \emph{arXiv Preprint},
  vol. abs/2112.12833, 2021. [Online]. Available:
  \url{https://arxiv.org/abs/2112.12833}
\BIBentrySTDinterwordspacing

\bibitem{DiBlase21}
G.~{Di Biase}, H.~Blum, R.~Siegwart, and C.~Cadena, ``{Pixel-Wise Anomaly
  Detection in Complex Driving Scenes},'' in \emph{Conference on Computer
  Vision and Pattern Recognition}, June 2021.

\bibitem{Hendrycks17b}
D.~Hendrycks and K.~Gimpel, ``{A Baseline for Detecting Misclassified and
  Out-Of-Distribution Examples in Neural Networks},'' in \emph{International
  Conference on Learning Representations}, 2017.

\bibitem{Jung21}
S.~Jung, J.~Lee, D.~Gwak, S.~Choi, and J.~Choo, ``{Standardized Max Logits: A
  Simple yet Effective Approach for Identifying Unexpected Road Obstacles in
  Urban-Scene Segmentation},'' in \emph{International Conference on Computer
  Vision}, 2021.

\bibitem{GluonCV20}
\BIBentryALTinterwordspacing
J.~Guo, H.~He, T.~He, L.~Lausen, M.~Li, H.~Lin, X.~Shi, C.~Wang, J.~Xie,
  S.~Zha, A.~Zhang, H.~Zhang, Z.~Zhang, Z.~Zhang, S.~Zheng, and Y.~Zhu,
  ``Gluoncv and gluonnlp: Deep learning in computer vision and natural language
  processing,'' \emph{Journal of Machine Learning Research}, vol.~21, no.~23,
  pp. 1--7, 2020. [Online]. Available:
  \url{http://jmlr.org/papers/v21/19-429.html}
\BIBentrySTDinterwordspacing

\bibitem{Yu19}
J.~Yu, Z.~Lin, J.~Yang, X.~Shen, X.~Lu, and T.~S. Huang, ``{Free-Form Image
  Inpainting with Gated Convolution},'' in \emph{Conference on Computer Vision
  and Pattern Recognition}, 2019.

\bibitem{Zhou17h}
B.~Zhou, A.~Lapedriza, A.~Khosla, A.~Oliva, and A.~Torralba, ``{Places: A 10
  Million Image Database for Scene Recognition},'' \emph{IEEE Transactions on
  Pattern Analysis and Machine Intelligence}, 2017.

\bibitem{Xie17b}
S.~Xie, R.~Girshick, P.~Doll{\'a}r, Z.~Tu, and K.~He, ``{Aggregated Residual
  Transformations for Deep Neural Networks},'' in \emph{Conference on Computer
  Vision and Pattern Recognition}, 2017.

\bibitem{Chen17g}
L.~Chen, G.~Papandreou, I.Kokkinos, K.~Murphy, and A.~L. Yuille, ``{Deeplab:
  Semantic Image Segmentation with Deep Convolutional Nets, Atrous Convolution,
  and Fully Connected CRFs},'' \emph{IEEE Transactions on Pattern Analysis and
  Machine Intelligence}, 2018.

\bibitem{Mukhoti18}
J.~Mukhoti and Y.~Gal, ``{Evaluating Bayesian Deep Learning Methods for
  Semantic Segmentation},'' in \emph{arXiv Preprint}, 2018.

\bibitem{Chan21b}
R.~Chan, K.~Lis, S.~Uhlemeyer, H.~Blum, S.~Honari, R.~Siegwart, P.~Fua,
  M.~Salzmann, and M.~Rottmann, ``{SegmentMeIfYouCan: A Benchmark for Anomaly
  Segmentation},'' in \emph{Advances in Neural Information Processing Systems},
  2021.

\bibitem{Lis23a}
K.~Lis, S.~Honari, P.~Fua, and M.~Salzmann, ``{Perspective Aware Road Obstacle
  Detection},'' \emph{IEEE Robotics and Automation Letters}, 2023, in Press.

\bibitem{PytorchResnext}
``{ResNeXt implementation in PyTorch Vision},''
  \url{https://pytorch.org/vision/stable/_modules/torchvision/models/resnet.html\#resnext101_32x8d},
  version 0.10.0.

\bibitem{PyTorch19}
\BIBentryALTinterwordspacing
A.~Paszke, S.~Gross, F.~Massa, A.~Lerer, J.~Bradbury, G.~Chanan, T.~Killeen,
  Z.~Lin, N.~Gimelshein, L.~Antiga, A.~Desmaison, A.~Kopf, E.~Yang, Z.~DeVito,
  M.~Raison, A.~Tejani, S.~Chilamkurthy, B.~Steiner, L.~Fang, J.~Bai, and
  S.~Chintala, ``{PyTorch: An Imperative Style, High-Performance Deep Learning
  Library},'' in \emph{Advances in Neural Information Processing Systems},
  H.~Wallach, H.~Larochelle, A.~Beygelzimer, F.~d\textquotesingle
  Alch\'{e}-Buc, E.~Fox, and R.~Garnett, Eds., 2019, pp. 8024--8035. [Online].
  Available:
  \url{http://papers.neurips.cc/paper/9015-pytorch-an-imperative-style-high-performance-deep-learning-library.pdf}
\BIBentrySTDinterwordspacing

\bibitem{Li18b}
W.~Li, O.~H. Jafari, and C.~Rother, ``{Deep Object Co-Segmentation},'' in
  \emph{Asian Conference on Computer Vision}, 2018.

\bibitem{Klambauer17}
G.~Klambauer, T.~Unterthiner, A.~Mayr, and S.~Hochreiter, ``{Self-Normalizing
  Neural Networks},'' in \emph{Advances in Neural Information Processing
  Systems}, 2017.

\bibitem{Kingma15a}
D.~P. Kingma and J.~Ba, ``{Adam: {A} Method for Stochastic Optimisation},'' in
  \emph{International Conference on Learning Representations}, 2015.

\end{thebibliography}
}

\clearpage
\appendices

\section{Implementation details}
\label{sec:sup_discrepancy network}

In this section, we present details on the discrepancy network architecture and training.

\subsection{Discrepancy Network Architecture}

The architecture is shown in Fig.~\ref{fig:sup_discrepancy_architecture}.
The network has two input streams: the original image and the image where the road area has been inpainted.
We use the pretrained ResNeXt101 network of~\cite{Xie17b} to extract features from both images.
We use the PyTorch implementation~\cite{PytorchResnext,PyTorch19} of this backbone and take the outputs of layers labeled \texttt{relu}, \texttt{layer1}, \texttt{layer2}, \texttt{layer3}.
At four levels of the feature pyramid, we fuse the two streams of features in these two parallel ways:
\begin{itemize}
	\item Concatenate stream 1 and 2, followed by a $1 \times 1$ convolution,
	\item Calculate pixel-wise correlation of features, following~\cite{Li18b}.
\end{itemize}
The results of the above are concatenated and passed on to an up-convolution pyramid 
which uses the SeLU~\cite{Klambauer17} activation function.
In the final step, the discrepancy score is multiplied by the binary drivable space mask,
since the outputs are only valid within the road area.

\subsection{Discrepancy Network Training}

The discrepancy network was trained for 65 epochs.
Each epoch iterates over the 2975 frames of our synthetic training set.
The training is done using $768 \times 384$ crops of the road area.
To improve training reproducibility, we pre-define the crops and their ordering in each epoch, and train all variants of the discrepancy network with the same sequence of samples.

We use binary cross entropy loss and the Adam~\cite{Kingma15a} optimizer.
We set the initial learning rate to $10^{-4}$ and then adjust it dynamically,
if there is no improvement of validation loss for 5 consecutive epochs, the learning rate is reduced 10 times.
We generate the validation set from Cityscapes validation subset in the same way as the training set.

\section{Lost\&Found region of interest}

When using the {\it Lost\&Found} dataset~\cite{Pinggera16}, 
we constrain the evaluation to the area of the ground truth free-space label,
and exlude the rest of the image.
This region is visualized in Fig.~7 in the main article.

\begin{figure} 
\centering
\includegraphics[width=\linewidth]{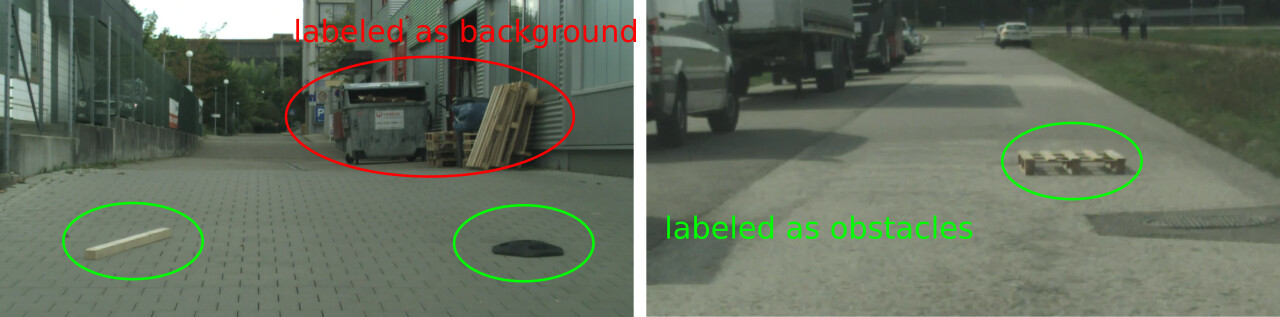}
\caption{In {\it Lost\&Found} only the central obstacles are labeled.}
\label{fig:laf_clutter}
\end{figure}

\begin{figure*} 
\centering
\includegraphics[width=0.83\linewidth]{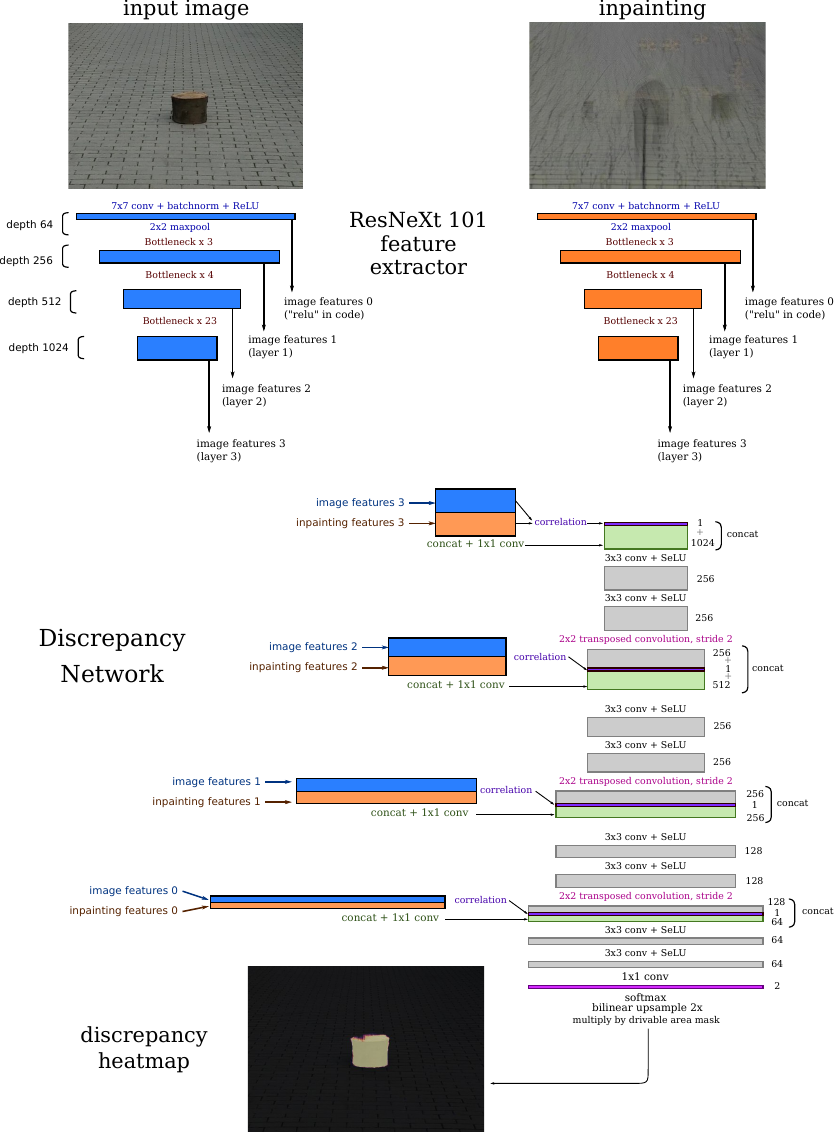}

\caption{\textbf{Discrepancy network architecture.}
}
\label{fig:sup_discrepancy_architecture}
\end{figure*}

While it would be interesting to test the detection of other clutter objects visible in the dataset,
the obstacles outside of the aforementioned free-space are not labeled.
Only the central objects are labeled, leaving the off-center obstacles (trash cans, pallets, fences, poles) marked as background. For example, in Fig.~\ref{fig:laf_clutter}, a wooden pallet is labeled as an obstacle when in front of the car but as background when leaning against a wall.

\end{document}